# EXPLORING PARKINSON'S DIAGNOSIS USING MACHINE LEARNING TECHNIQUES

MASTER'S PROJECT – F21MP

Author
***Farhaan Qazi***

Supervisor
***Dr. Alistair McConnell***

Co-Supervisor
***Dr. Marta Vallejo***

HERIOT WATT UNIVERSITY

**AUGUST 20, 2023**
**MACS, HERIOT WATT**
**RICCARTON, EDINBURGH**

## *ABSTRACT*

In recent years, machine learning (ML) techniques have emerged as promising tools for accurate and timely diagnosis of Parkinson's disease (PD) which is an incurable neurodegenerative disorder. This study explores the integration of ML algorithms and PD diagnosis, highlighting their potential applications in various aspects of the disease. ML methods such as support vector machines, decision trees, multilayer perceptron, *k*-nearest neighbours, logistic regression and random forest (with and without particle swarm optimisation) were applied on a PD dataset obtained from Leeds Teaching Hospitals NHS Trust containing data from 58 people. In terms of accuracy, the Decision Tree and Logistic Regression performed the best with scores of 72-76% while multilayer perceptron was least effective. All the models were remarkable in classifying positive PD cases with sensitivity scores going up to 97.56%, with SVM classifier at the pinnacle. However, ability to classify healthy controls from PD cases was insignificant throughout the board. Although, all the employed models delivered reasonable results in categorising PD patients in the presented dataset; access to more data, further study and advancement of these ML approaches could lead to enhanced diagnostic outcomes, supplementing to an improved understanding of Parkinson's disease.

"Parkinson's has taught me the importance of patience, perseverance, and acceptance."

*Muhammad Ali*

# TABLE OF CONTENTS

## *List of Figures –*

## *List of Tables –*

## *Abbreviations*

PD – Parkinson's Disease
NHS – National Health Services
ML – Machine Learning
RNN – Recurrent Neural Network
CNN – Convolutional Neural Network
SVM – Support Vector Machines
XGBoost – Extreme Gradient Boost
FS – Feature selection
SWEDD – Scans Without Evidence of Dopaminergic Deficit
HC – Healthy Control
RF – Random Forest

Word Count – 18268 words

# *Chapter – 1* *INTRODUCTION*

Parkinson's disease (PD) is a degenerative movement disorder that primarily affects the human nervous system. It is sparked by a deficiency of dopamine which is responsible for control of movements in the body. This deficiency is caused by the death of dopaminergic neurons in the brain which is supplemented by aberrant protein deposits in the brain known as Lewy bodies, and the subsequent inflammation of brain neurons (Nour et al., 2023). These aberrant processes collectively alter brain chemicals, causing difficulties in essential human brain functions such as thinking, movement, behaviour, and mood. PD is characterised mostly by tremors (unconscious shaking of the limbs), bradykinesia (slow limb movements), muscle inflexibility, motor dysfunction, diminished agility, a stooped posture, and a distorted facial expression. Non-motor symptoms such as depression, loss of motivation, and anxiety have also been observed in PD patients, but the loss of motor skills over time is more generic to PD (Jellinger, 2003). Due to its chronic nature, this disease is permanent and the underlying cause of the abnormalities triggering the PD is still unknown.

At present, there is no cure available for PD but treatments exist that can reduce the severity of symptoms and improve the quality of life (Mei et al., 2021). Patients with PD often have non-motor symptoms long before they have motor symptoms. However, in order to make an accurate clinical diagnosis and choose the right treatment procedure, the treatment is based on the motor symptoms, since the non-motor symptoms are often hard to spot and are often attributed to other diseases and age (Pan et al., 2012). It is a common medical practice to take all the aspects of human conditions into account like medical history, examinations, tests, mental tasks, etc to conclude on PD because no specific test has been developed for PD so far due to the lack of a biomarker. Medical fraternity usually involve advanced medical assessments on PD patients to observe the brain activity but only to rule out other medical conditions (Goetz, 2011).

The advent of technology with its irresistible influence on human lives in the latter half of the 20th century exhibited an immense potential to aid the exploration of novel approaches and methods of treatments which manifested the onset of technological aspects such as artificial intelligence (AI) and machine learning (ML) into the medical sector (Naqvi, 2023). ML is a subfield of AI that enables computers to acquire knowledge without being explicitly programmed. In medical research, ML can be utilised to analyse enormous quantities of medical data in order to uncover patterns and connections that may be invisible to the human eye. This knowledge can then be used to build novel diagnostic tools, treatments, and preventative measures (LeCun et al., 2015).

## *1.1 Motivation*

Despite intensive research efforts, the causes and mechanisms of PD remain unknown, impeding the development of effective treatments and preventative measures. So, a fresh approach such as ML with incredible capabilities was inevitable. ML has emerged as an effective technique that can potentially upgrade the understanding of PD. By utilising the capabilities of this field, hidden patterns can be found, early-stage symptoms can be identified, and disease progression can be anticipated. All of which can ultimately contribute to the creation of revolutionary therapeutic options (Emamzadeh and Surguchov, 2018). Several ML techniques have been adopted and other ML algorithm-based models are being introduced to improve the diagnostic accuracy of PD. Modern ML techniques have aided in the early detection of PD by training ML algorithms on patient data.

Further, there is a high rate of clinical misdiagnosis of PD, approximately 10-17% and it usually takes around 3 years to reach a 90% diagnostic accuracy (Hughes et al., 2002). So, since PD has such a profound effect on people all over the world, the author felt motivated to deeply explore the disease and its diagnosis using the knowledge of ML and was driven by a desire to make a difference in the course of PD by focusing on raising its understanding.

## 1.2 Aim

In this dissertation, the author aimed to –

- Review and analyse the existing literature on the application of ML techniques for PD diagnosis to investigate its effectiveness to measure and predict PD symptoms by leveraging the already available data.
- From the existing research, evaluate and summarise the advancements and achievements in the field of ML-based PD diagnosis to identify gaps and areas for further research and development.
- Analyse the performance of common independent ML models such as Decision Trees, MLP, KNN, Random Forest, SVM, Logistic Regression and others to the data set acquired from Leeds Teaching Hospitals NHS Trust and compare the effectiveness, accuracy, and reliability of these algorithms.

By addressing the above-mentioned aims, the author in this paper intended to provide insights into the potential and limitations of the mentioned ML techniques as a diagnostic tool for PD.

## 1.3 Report Outline

The author saw it fitting to organise this dissertation in such a way that the *introduction* chapter provides an outline to the chosen topic of research. Then, the *literature review* chapter analyses and studies the existing literature on the subject matter. It is followed by the *requirement analysis* chapter which is an account for the requirements of this dissertation along with their priorities. Beyond, the *ethics* chapter provides the ethical, legal and other aspects to consider while carrying out the research. Further, the *implementation* chapter describes in detail the data and the various methods used, algorithms applied, and the experiments performed. Consequent, the *evaluation and results* chapter evaluates all the findings from the wide experiments completed. Finally, the *discussion and future work* chapter closes the dissertation with appropriate comparisons, a conclusion and future work.

# Chapter – 2 LITERATURE REVIEW

## 2.1 Literature Search Strategy

A thorough search approach was employed to find the appropriate existing literature on the subject because the robustness of the search is closely connected with the validity of a review (Dignen, 2008).

### 2.1.1 Databases

After making a list of the appropriate databases, the author chose the academic citation databases that are the most frequently opted for and were most suitable for this research. To be certain of the authenticity of the papers, the author obtained most of the papers from PubMed and Medline ProQuest. These databases are a valuable and commonly sought sources for biomedical publications and health care interventions (Lu, 2011).

The author also used other recognised databases such as IEEE Xplore and ResearchGate which are an abode of technical literature. These are a go-to reference for engineering research of all kinds. It includes millions of full-text journal articles from all disciplines of engineering ("A-Z Databases: Computer Science; IEEE,").

### 2.1.2 Keywords

The author picked the most relevant keywords for the search in order to guarantee the incorporation of studies that provide answers to the concerns raised in the research. Primary keywords used to search for the research papers were "parkinson's disease", "diagnosis", "clinical methods" "technologies", "techniques", "methods", "machine learning", "algorithms", "SVM", "CNN", "k-means" "ensemble", "neural nets". Most of the queries submitted to the databases were of the form – "parkinson's disease" AND "diagnosis" AND "technologies" OR "methods" OR "machine learning". These keywords were used across all the databases.

### 2.1.3 Screening

To extract pertinent material, the author relied primarily on academic papers that had been reviewed by other academics and were written in English. In addition, the author had to consult grey literature also, by using "google scholar" which provided a quick way to search through a huge number of academic publications (Lu, 2011).

## 2.2 PD and the cure

At present, there is no cure available for PD but treatments exist that can reduce the severity of symptoms and improve the quality of life (Mei et al., 2021). Patients with PD often have non-motor symptoms long before they have motor symptoms. However, in order to make an accurate clinical diagnosis and choose the right treatment procedure, the treatment is based on the motor symptoms, since the non-motor symptoms are often hard to spot and are often attributed to other diseases and age (Pan et al., 2012). It is a common medical practice to take all the aspects of human conditions into account like medical history, examinations, tests, mental tasks, etc to conclude on PD because no specific test has been developed for PD so far due to the lack of a biomarker. Medical fraternity usually involve advanced medical assessments on PD patients to observe the brain activity but only to rule out other medical conditions (Goetz, 2011).

The advent of technology with its irresistible influence on human lives in the latter half of the 20th century exhibited an immense potential to aid the exploration of novel approaches and methods of treatments which manifested the introduction of artificial intelligence (AI) and machine learning (ML) into the medical sector (Naqvi, 2023). ML is a subfield of AI that enables computers to acquire knowledge without being explicitly programmed. In medical research, ML can be utilised to analyse enormous quantities of medical data in order to uncover patterns and connections that may be invisible to the human eye. This knowledge can then be used to build novel diagnostic tools, treatments, and preventative measures (LeCun et al., 2015).

## 2.3 Key symptoms of PD

From tremors and change in gait to issues with sleep and being anxious all the time, PD patients can experience more than 40 different symptoms (Pan et al., 2012). PD individuals might experience these symptoms in a variety of distinct ways, and the course of the disease for each person can be entirely unique and can change even by an hour. Symptoms such as tremor, rigidity (stiffness), slowness of movement, minor memory loss, thinking problems, sleep problems, pain, and mental health issues such as anxiety and depression are a few of the most typically observed signs of PD (Hausdorff et al., 2000).

The most common symptoms experienced by PD patients are linked to motor abilities but there can be non-motor symptoms as well. Since this study focused primarily on drawing data taken from hand-tasks completed by PD patients, the author wanted to emphasise on motor symptoms of PD, particularly tremors which are most likely to affect patients the most, while performing the aforementioned tasks.

### 2.3.1 Tremor

A tremor is when a part of the body shakes on its own (involuntary), usually the hand. Along with moving slowly and being stiff, tremor is one of the most noticeable signs of PD. Because everyone is different, having a tremor doesn't always mean that you have PD. It can also be a sign of some other diseases ("Parkinson's symptoms," n.d.). There are majorly 2 types of tremors which can be identified while studying this disease. First, *essential tremor* is a shaking of the hands, head, legs, body, or voice that is most noticeable when a person is moving. It is commonly mistaken for PD. Second, *dystonic tremor* may develop in patients who suffer from a range of movement disorders that cause muscle spasms and contractions called dystonia. It is usually quiet challenging to distinguish between these tremors from a PD tremor. Data scientists who are currently working on PD have one of their main objective to over this challenge of pointing out the PD symptoms from other similar symptoms.

Patients diagnosed with PD can show 2 types of tremors –

- *Resting tremor* – they can happen when the person is calm, still, or even sleeping.
- *Action tremor* – they can happen only when the person is moving or in action.

(Gallicchio et al., 2018; Nour et al., 2023)

### 2.3.2 Bradykinesia

Bradykinesia is characterised by slowing of all voluntary movements, as well as a steady diminution in size and velocity as the movement proceeds. Bradykinesia is fundamental to the definition of parkinsonism and the diagnosis of PD and atypical parkinsonism (AP). Current diagnostic criteria for PD include both slowness and cumulative reduction in movement amplitude and velocity when performing repetitive motions such as decrement or sequence effect, as diagnostic criteria for bradykinesia. Evaluation scales often used in clinical practise represent the current understanding of bradykinesia, which encompasses possibly diverse motor disorders (Shrimanker et al., 2023).

❖ The author mainly focused on tremor and Bradykinesia because the International Parkinson and Movement Disorder Society has provided a set of criteria for PD diagnosis that could be considered a modernised version of the decades-old Queen's Square Brain Bank Criteria. These criteria are based on a professional neurological examination that confirms the presence of bradykinesia and at least one additional cardinal motor symptom, such as stiffness or resting tremor, to confirm PD in a person (Tolosa et al., 2021).

## 2.4 Clinical Diagnostics of PD

As there are no specific laboratory tests that can diagnose PD, clinical diagnosis is always a challenge. The clinical process commonly starts by a neurologist-based assessment of the medical history of patient and the family trailed by a physical exam, and a few basic tests. Although some advanced tests may be done to rule out other conditions that can cause similar symptoms (Orimo, 2013). The physical exam concludes by making eye-judgements of tremors, stiffness, gait, slow movement of the limbs and the body as a whole where the posture, balance, and coordination is also checked by the doctor. Afterwards, specific brain imaging techniques such as CT Scan, MRI Scan, and SPECT Scan, are used to evaluate the brain function (bgdteam, 2022).

Some of the common clinical methods for identifying PD are –

- Unified Parkinson's Disease Rating Scale (UPDRS) –

A standardised scale used to evaluate the severity of Parkinson's disease symptoms. It was created by a group of neurologists and researchers in 1987 and has been changed numerous times since then. UPDRS is a reliable and effective assessment technique for Parkinson's disease symptoms. It is frequently utilised in clinical trials and research investigations to assess the efficacy of new medicines. The UPDRS can also be used to monitor the progression of the disease over time and to assist with treatment decisions (DeMaagd and Philip, 2015).

- Hoehn and Yahr scale –

Margaret Hoehn and Melvin Yahr published in 1967 a staging method used to classify the course of PD. It is an easy-to-use and straightforward tool for tracking the progression of the disease. It is frequently employed to aid with treatment decisions and future planning (DeMaagd and Philip, 2015).

- DaTSCAN –

A brain imaging test that can help to distinguish PD from other movement disorders. It's a specialised imaging technique which is gaining popularity and has been frequently utilised by the medical community to diagnose PD. In this test, a scanner detects a dye put into the patient's bloodstream during this imaging procedure. The scan can then detect the reduction of dopamine transporter in the brain. With a sensitivity of 98% and a specificity of 67%, DaTSCAN has proven to be highly useful in early PD diagnosis (de la Fuente-Fernández, 2012). In the early stages of PD, the clinical diagnosis is 84 percent accurate, while in the later stages, it is 98 percent correct (de la Fuente-Fernández, 2012).

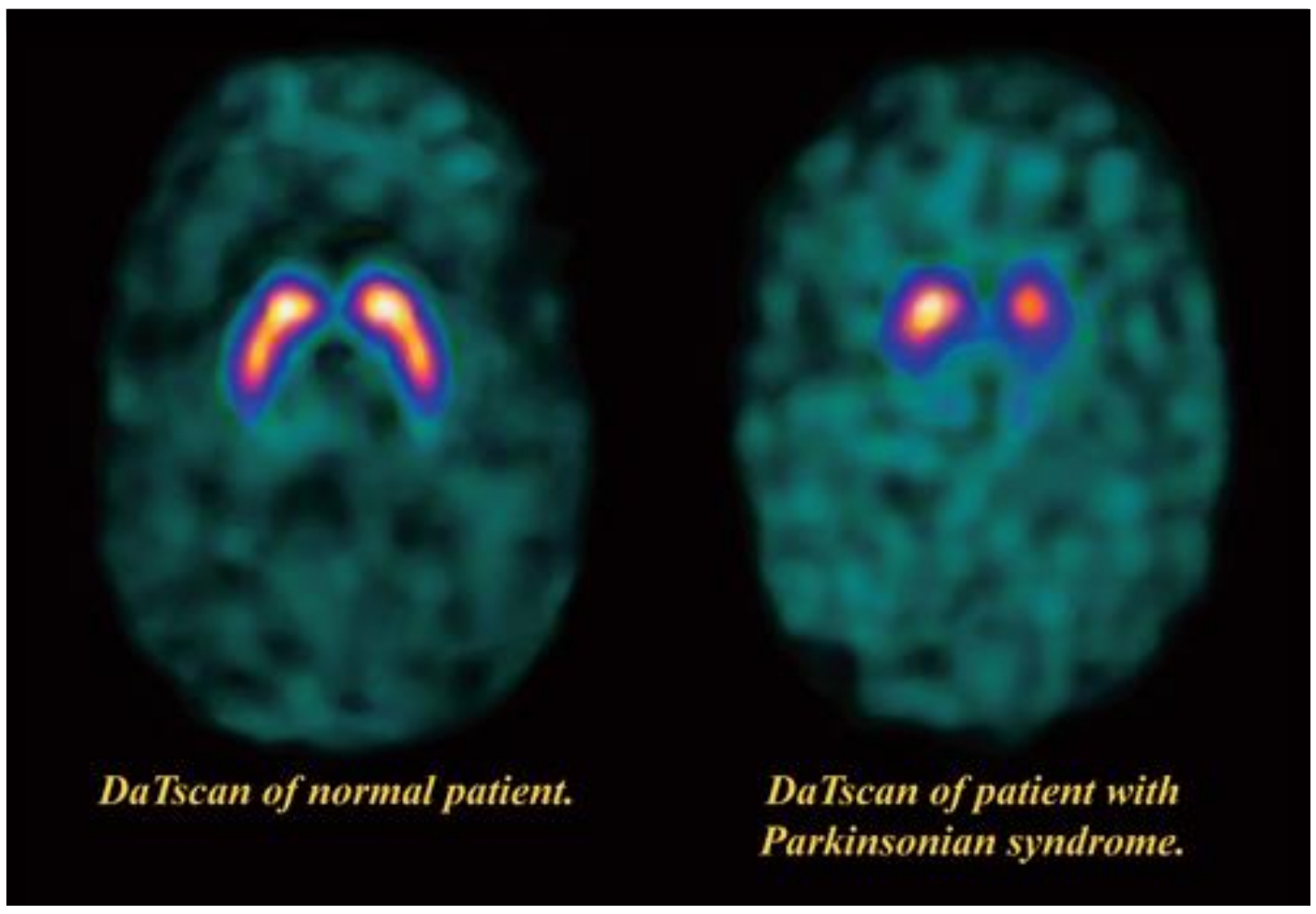

Fig 2.1 Brain imagery from DatSCAN [Image source – (de la Fuente-Fernández, 2012)

- Genetic testing –

It is a relatively recent diagnostic and management tool for PD. It can be used to identify those who are at a higher risk of developing Parkinson's disease, even if they do not currently exhibit symptoms. This information can be utilised to make informed choices on lifestyle modifications and early treatment. A certain types of genes have been identified to be predominantly found in the people having PD than general population. So, the doctors may recommend genetic testing to a person having a family history of PD to check for these particular genes (DeMaagd and Philip, 2015).

- Biomarkers –

Biomarkers are chemicals that can be tested in the blood or other tissues to suggest the presence of PD. They are quantitative indicators of biological processes or infectious diseases. Biomarkers can be utilised in the context of PD to diagnose the disease, track its course, and predict how patients will respond to treatment (DeMaagd and Philip, 2015).

From the above discussion of clinical ways of diagnosis of PD, the author came to believe that PD diagnosis can be demanding, but simultaneously an early detection can be assuring of patients receiving the most effective treatment. The techniques discussed by the author in this section have the potential to increase the accuracy of PD diagnoses and can lead to the development of new treatments for the disease.

## *2.5 Difficulties in Clinical Diagnostics*

PD has been an incurable condition for such a long period of time, it has sparked disputes both inside and outside of the medical community. And the most frequent of the disputes has undoubtedly been its misdiagnosis. In the recent medical history, the greatest challenge by far has been differentiating PD from a variety of neurodegenerative disorders and most commonly has been confused with essential tremor and secondary parkinsonism (Chen et al., 2015).

A recent meta-analysis has revealed that only 86.6% of 11 clinicopathological examinations showed a diagnostic accuracy for the clinical diagnosis of PD (Giovanni Rizzo et al., 2016). Further, it has been observed that at least 10% of patients who were diagnosed by the doctors of having PD, had other illnesses with similar symptoms even when strict clinical diagnostic standards were being applied. The main reason behind the incorrect diagnose in these cases is that the symptoms of other conditions are similar to those of PD. In addition, error rates range from 15% to 24% across series, indicating widespread misclassification (Tolosa et al., 2021).

## 2.6 PD and the world

An estimated 10 million people afflicted worldwide and 145,000 in the UK alone makes the severity of the incurable disease quite evident. In addition, the numbers are expected to rise to 13 million by 2040 is a cause for rising concern (Naqvi, 2023).

Table 2.1 – PD Statistics

| Description | Value |
|---|---|
| PD worldwide | 10.6 million |
| PD patients added each year | 60,000 |
| Average age of onset | 60 years |
| Male to female ratio | 1.3 : 1 |
| PD in the United Kingdom | 1.3 % |
| PD in the United States | 1.4% |
| PD in Europe | 1.3 - 1.8 % |
| PD in Asia | 1.2 - 1.6 % |
| Life expectancy after diagnosis | 10-15 years |
| Leading cause of death in PD patients | Respiratory failure |

("World Health Organization (WHO)," 2023, "Parkinson disease," 2023)

PD disproportionately affects the elderly, and the risk increases with age. It is more prevalent in people over the age of 60, but younger people can still be affected. It is responsible for a considerable number of disability life years and deaths on a global scale, resulting in an extraordinarily high demand for essential health resources. The disorder affects around 1% of the world's population, or approximately 10 million people ("What is Parkinson's? | Parkinson's Europe," 2020, "World Health Organization (WHO)," 2023). As the incidence of PD rises with age and as people live longer owing to better living and health standards, the disease's prevalence is projected to expand considerably in the future. According to the 2015 Global Burden of Disease Study, by 2040 almost 13 million people could have PD ("Global, regional, and national burden of Parkinson's disease, 1990–2016: a systematic analysis for the Global Burden of Disease Study 2016 - The Lancet Neurology," n.d.). An estimates approximates that 1 in 70 people will be diagnosed with PD in the UK during their lifetime ("Media and press office," 2023).

## 2.7 Need for novel diagnostics methods

As the author discussed before that clinical diagnostics of PD can be difficult since symptoms sometimes overlap with those of other illnesses, making an accurate diagnosis problematic or impossible. Even when precise diagnostic criteria are applied, a sizeable number of patients initially diagnosed with PD turn out to have other disorders with similar symptoms. Such factors have always emphasised the need for additional and latest approaches, such as ML to enhance the classification and diagnostic accuracy of PD.

ML algorithms can understand patterns and associations that human therapists might miss if they just had access to a single dataset during training. Clinical notes, imaging results, genetic information, and biomarkers could all benefit from being analysed by ML algorithms. Misdiagnosis of PD can be reduced with the help of these models, which can shed light on the disease and other similar diseases (Guida and Sacco, 2019). The use of ML methods in the diagnosis of PD has already shown encouraging results. Neuroimaging data, including as MRI and DAT scans, can now be analysed by algorithms specifically designed to spot the small changes in the brain that are characteristic of PD. These algorithms can even distinguish PD from other illnesses with similar symptoms by identifying biomarkers that are particular to the disease (Guida and Sacco, 2019).

## 2.8 ML Techniques Used For PD Diagnosis

The importance of usage of ML algorithms in the diagnosis and treatment of PD has grown steadily since their introduction to this field. Since ML may aid in the analysis of large amounts of patient data faster and more reliably than the standard diagnostic approaches, it has gained increasing recognition in the medical community (Bologna et al., 2023). As mentioned in the previous discussions, diagnosing PD can be challenging because its symptoms are nonspecific and are often similar to other diseases. But these similar looking symptoms can be segregated by the application of ML algorithms which have the ability to identify patterns specific to PD by training on large patient datasets. These algorithms can also be used to construct predictive models that support the identification of patients at risk for developing PD (O'Shea and Nash, 2015).

As a result, physicians will be able to intervene sooner, perhaps pausing or reducing the progression of the disease. Depending on the requirements of a specific study or clinical application, the most prevalent ML algorithm utilised in PD diagnostics can vary.

To list a few –

- Support Vector Machines (SVM)
- Artificial Neural Networks (ANN)
- Convolutional Neural Networks (CNN)
- K-Nearest Neighbour (KNN)
- Hybrid methods
- Linear regression
- Logistic regression
- Decision trees
- Random forest
- Naïve bayes

### 2.8.1 SVM and PD Diagnostics

Support Vector Machines (SVM) is an ML technique which classifies new cases using non-probabilistic binary classifiers. It first analyses known data and then classes unknown test samples. The linear SVM classifier divides examples into two locations in space and maximizes the gap width. SVMs classify or regress using infinite-dimensional hyperplanes. Classifier error is lowest for a hyperplane with largest separation from nearby training data points. These hyperplanes are built by the help of the data points which are closest to the hyperplane, these datapoint are called Support Vectors.

SVM is usually executed in a linearly but when a satisfactory fit is not obtained, non-linear SVM performs better. Non-Linear SVM maximises hyperspace by mathematically transforming feature space (S. Shetty and Y. S. Rao, 2016).

SVM has become remarkably useful for PD diagnosis due to its high-dimensional data handling, robustness to overfitting, non-linear classification, robustness to outliers, well-established theory and implementation, and potential interpretability (Karapinar Senturk, 2020).

(S. Shetty and Y. S. Rao, 2016) considers an approved and previously verified data set containing 12 features related to human gait cycle. The data set consists of records from 64 patients obtained from 'The National Institutes of Health-sponsored Research Resource for complex physiological signals'(Hausdorff et al., 2000). SVM technique was applied to the data to classify PD and healthy patients.

During the pre-processing of data, Feature Selection (FS) which identifies the key features of the problem was employed. It was found there was a complete negative correlation among some features and were hence discarded to avoid having redundancies. Improving classification precision depends heavily on an accurate determination of characteristics and dimensionality reduction can boost the overall efficiency of ML techniques (Jain and Singh, 2018). A good correlation was observed between 10 features, so only those were used for the actual experiment. Before actually testing the data using the SVM, feature vectors were calculated using a host of statistical tools. Further reduction of the preliminary vectors was done by computing the SVM for each vector individually. The author of this paper developed the classifier model using MATLAB.

*Result* – The classifier achieved an accuracy of 83.33%. The model gave 75% true positive and 25% false negative outcomes.

*Evaluation* – Even though FS was used before applying the classifier to the dataset, accuracy was not up to the mark. The author discusses SVM further.

### 2.8.2 ANN and PD Diagnostics

Neural networks are ML techniques inspired by the operation of human brain. Similar to neurons in humans, neural networks have neurons each representing a feature from the data. These neurons pass the feature data points through the hidden layers (if present) using random weights and bias. The hidden layers assist in intense understanding of the patterns and relationships underlying the data. The network then makes the corresponding prediction and produces the result through an output later (LeCun et al., 2015).

(Karapinar Senturk, 2020) used a dataset created by Max Little with the cooperation of the National Voice and Speech Centre of the University of Colorado and the University of Oxford. The dataset was taken from the UCI – Machine Learning Repository ("UCI Machine Learning Repository: Parkinsons Data Set,"). The dataset consists of 195 biomedical sound measurements taken from 31 people consisting of 8 healthy subjects and 23 with PD.

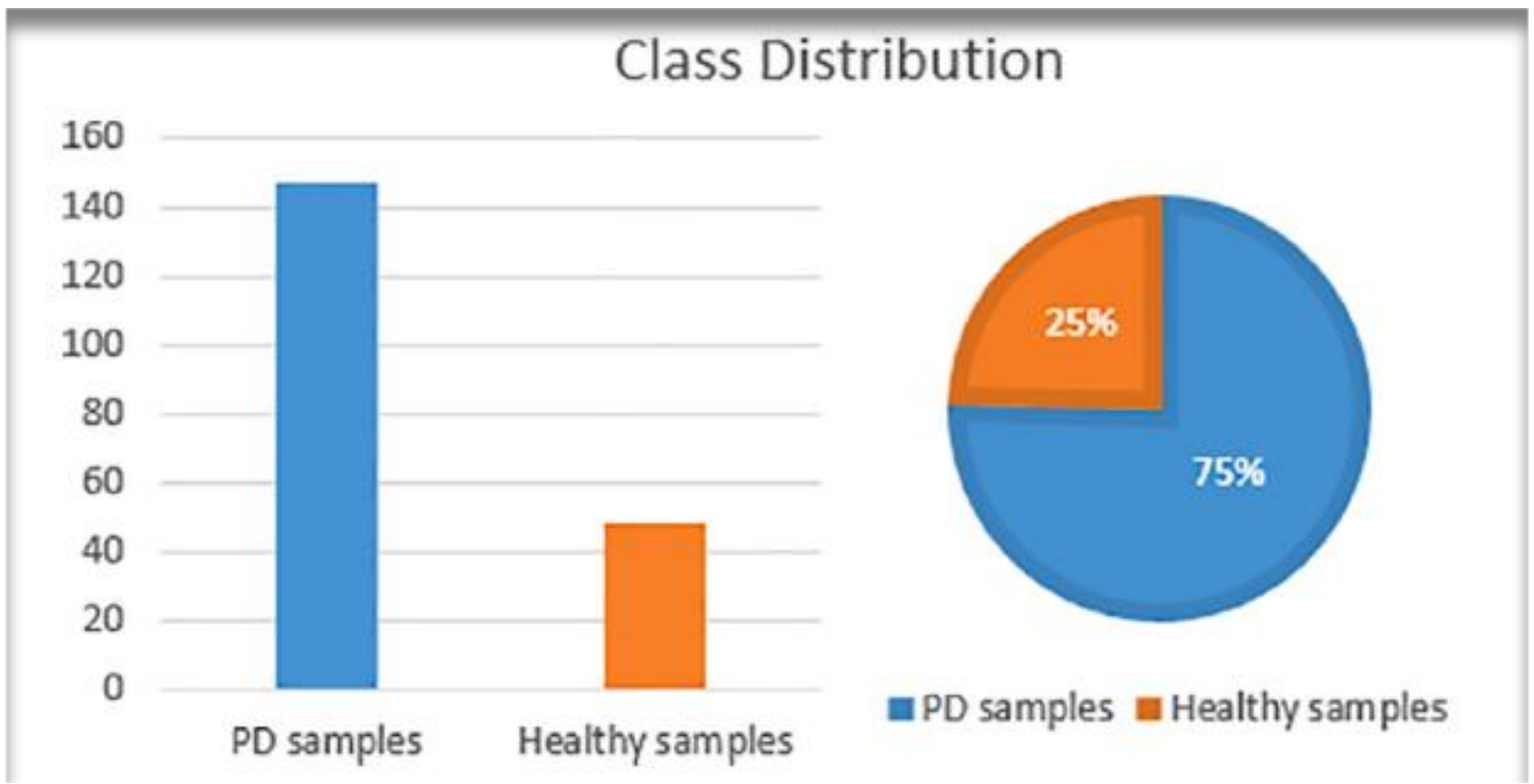

Fig 2.2 Class distribution in the dataset [Image source – (Karapinar Senturk, 2020)]

Three different classification algorithms were applied to this data – CART, SVM & ANN and the end results were compared based on multiple criteria.

*Classification and Regression Tree (CART)*

The CART method is a nonparametric approach in identifying the most significant independent and interactive variables. If the outcome is a continuous variable, CART makes regression trees. If the outcome is a categorical variable, CART makes classification trees. If the outcome is a continuous variable, CART makes regression trees. If the outcome is a categorical variable, CART makes classification trees (Yohannes and Webb, 1999).

Further, as good practice, during the pre-processing FS was done on 23 voice features and only effective features were used, reducing the analysis cost. Different FS methods were used such as Feature Importance (FI) method for CART and Recursive Feature Elimination (RFE) method for SVM and ANN. For an efficient classification, the dataset was rearranged so that there were fewer columns (features).

Also, interestingly the whole experiment was designed in such a way that while comparing the 3 different classifying methods, benefit of application of FS was also evaluated (discussed in the results). The experiments were performed using 7 features for CART and 13 features for SVM and ANN methods.

*Result* – Table 2.2

| ML techniques used | Accuracy without Feature Selection | Accuracy without Feature Selection |
|---|---|---|
| CART | 0.8523 | 0.9076 |
| SVM | 0.7998 | 0.9384 |
| ANN | 0.8025 | 0.9154 |

*Evaluation* – It is clear from the results of the experiment that although SVM with FS had the best performance at classifying (93.84%) but ANN also performed (91.54%) reasonable as well. Again, we have established the efficient application of SVM to PD diagnosis and simultaneously established the advantage of careful use of FS in data pre-processing. Further, from these two experiments it can be said that SVM classifier is fairly successful for PD diagnosis.

### 2.8.3 CNN and PD Diagnostic

The Convolutional Neural Network (CNN) is a well-known type of artificial neural networks (ANN) that is mostly used for image pattern recognition (O'Shea and Nash, 2015). CNNs display greater performance when presented with inputs such as images, speech, and audio signals, as a result of their specific operational approach. Consisting of three core layer types - Convolutional, Pooling, and Fully connected (FC) - CNNs operate in a distinctive manner. Initialized by the convolutional layer, the initial layer of the network concentrates on detecting complicated patterns. In spite of the possibility of future convolutional or pooling layers, the ultimate layer is the one with complete connectivity. Initial layers concentrate on elemental properties such as colours and edges. Subsequently, when the visual data passes the succeeding layers, the network gradually discerns more wide parts or outlines of the subject matter, ultimately achieving accurate recognition ("What are Convolutional Neural Networks," IBM).

(Gil-Martín et al., 2019) did a study where CNN was applied to a public dataset comprising of 77 unique spiral drawings from 15 healthy and 62 people diagnosed with PD. The CNN model with two convolutional layers having 16 filters and a maxpooling layer for main feature extraction followed by three fully connected layers for final classification. A dropout layer was also introduced to avoid over fitting. The inputs were compiled in a 2 D matrix with N*125 dimensions. As it was a binary classification, a sigmoid function was used as an activation function and binary cross-entropy as loss metric. The primary experiment employed 25 epochs, 100 batches, and ReLU as the activation function for the deep learning framework. Root-mean-square propagation was fixed as the optimiser.
*Result* – After employing a five-fold cross-validation to split the data into train & test data and repeating the experiment five times modifying the test and training sets each time, the system reported an accuracy of 96.5% and F1-score of 97.7%.

*Evaluation* – On comparing the results with similar previous experiments carried out on the same dataset, (Gallicchio et al., 2018) obtained an accuracy of 89.3% while (Khatamino et al., 2018) reported an accuracy of 72.5% for the subject-wise cross-validation. The authors believe the use of the spectrum points as inputs to the CNN instead of the raw data has wholly improved the output accuracy.

### *2.8.4 Multi-class SVM and PD Diagnostics*

A similar study was done by (Singh et al., 2016) where MRI data was fed into multi-class SVM for classification of PD and non-PD patients. The data was obtained from Parkinson's Progression Markers Initiative (PPMI, a public repository) and contained 150 MRIs, 50 each for healthy control (HC), *de novo* PD and Scans Without Evidence of Dopaminergic Deficit (SWEDD) subjects.
FS method called principal component analysis (PCA) was employed for feature extraction and reduction in data-dimensionality from the high-dimensional imaging data. The ability of a diagnostic test to differentiate between healthy and diseased patients is known as its *discriminative ability* and the discriminative ability of a diagnostic procedure is called diagnostic accuracy ("Feature Importance," codeacademy). After arranging the features in descending order based on their collective FDR scores, a binary and multi-class SVM was applied to the MRI (image) data.

The authors utilised a 10-fold cross-validation for evaluation of the performance of the methodology which was based on the accuracy for every comparison identifying the grey matter (GM) & the white matter (WM) in the brain.

*Result* –

Overall mean accuracy for *binary classification* was approximately 99%, 96% and 96% for GM brain images and approximately 88%, 94% and 94% for WM. Overall mean accuracy for *multi-class classification* were 86% and 87% for GM & WM respectively.

*Evaluation* –

Evaluating the performance of the methodology based on the accuracy, a similar study was done by (Duchesne et al., 2009) where an accuracy of 91% was achieved, which is relatively lower than (Singh et al., 2016).

## 2.8.5 Ensemble Models

### 2.8.5.1 RF and XGBoost

Several individual decision trees working together to provide a collective decision on a specific task is a *random forest* (RF). Each tree in an RF produces its own outcome and the most recurring outcome from all the decision trees for a particular prediction eventually develops into the final conclusion of the RF model (Yiu, 2021).

*Extreme Gradient Boosting* (XGBoost) is a supervised tree-based ML algorithm. It includes a scalable and distributed gradient-boosted decision tree that offers concurrent tree boosting and is an important ML library for regression, classification, and ranking issues (Chen et al., 2015).

(Xing et al., 2022) conducted an experiment on 398 patients with confirmed upper limb tremors. The data was obtained from department of Neurology of Rui Jin Hospital (Shanghai, China). RF classifier and XGBoost among others were applied in combination for classification of either PD or essential tremor (ET). For training, forty tremor variables and two demographics (sex and age) were selected during data preprocessing. Grid-search approach was employed for hyperparameter selection and tuning before splitting (80/20) the data into a training and test samples, followed by a 10-fold cross-validation on the training sample.

*Result* –

RF and XGBoost performed reasonably with an accuracy rate of 84%.

### 2.8.6 Hybrid Models

(Varalakshmi et al., 2022) in their research developed several predictive models to determine the best ML technique for classifying PD in the presented dataset obtained from Kaggle repository. Dataset comprised of spiral hand drawing data from 51 healthy controls and 51 confirmed PD subjects. They used a few simple ML models before divulging into intricate deep learning (DL) models and further into hybrid models.

As suggested by the name, *hybrid models* can be created by developing two or more models together to produce fresh models depending on the requirement of the problem. Hybrid models can be designed by merging DL networks with ML frameworks or DL with DL models (Drotár et al., 2016).

*Results* –

At the end of the experiment, following results were observed –

Table 2.3 –

| Model type | Model with highest accuracy | Accuracy (%) |
|---|---|---|
| ML | SVM | 82.40 |
| DL | RESNET50 | 96.79 |
| Hybrid (ML + DL) | RESNET50 + SVM | 98.45 |
| Hybrid (DL + DL) | RESNET + MLP | 98.36 |

## 2.9 Models studied before the experiments

Here, the author has elaborated a few ML models that were studied before beginning the development of the prototype for PD diagnosis.

Based on the requirements of the experiment, all ML models are categorised as supervised and unsupervised. Supervised models are further divided into regression and classification models (Shin, 2022). In supervised learning, a function is created which learns by analysing pairs of inputs and outputs and accordingly maps a new unseen input to an output.

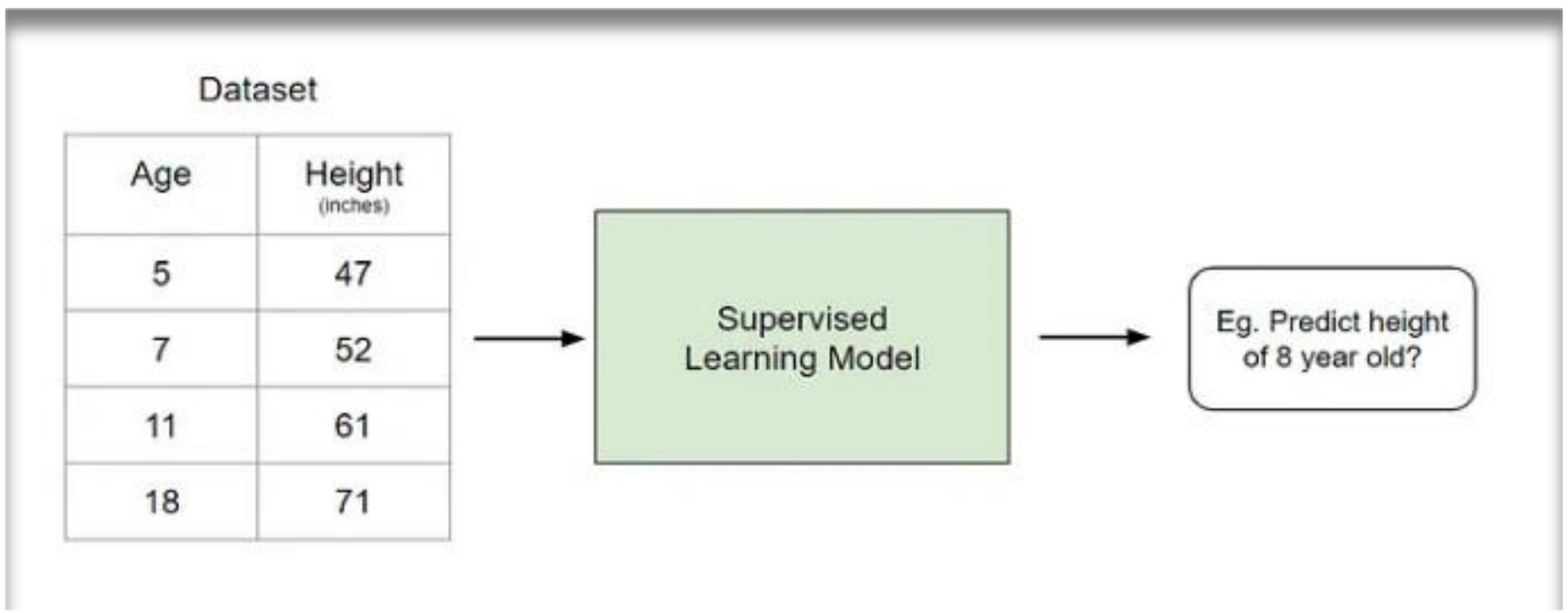

Fig 2.3 Example of working of supervised learning [Image source – (Shin, 2022)]

From supervised learning, the author examined –

### 2.9.1 Linear Regression

Linear regression is a frequent and straightforward technique for predicting numerical values based on input information (Lahmiri et al., 2018). It works under the assumption that there is a linear connection between the data being input and the variable being sought for, and it searches for the line that provides the greatest fit while simultaneously reducing the amount of error that is left behind. Widespread applications for linear regression include predicting housing prices, clinical data, stock prices, and other numerical quantities (Rana et al., 2022).

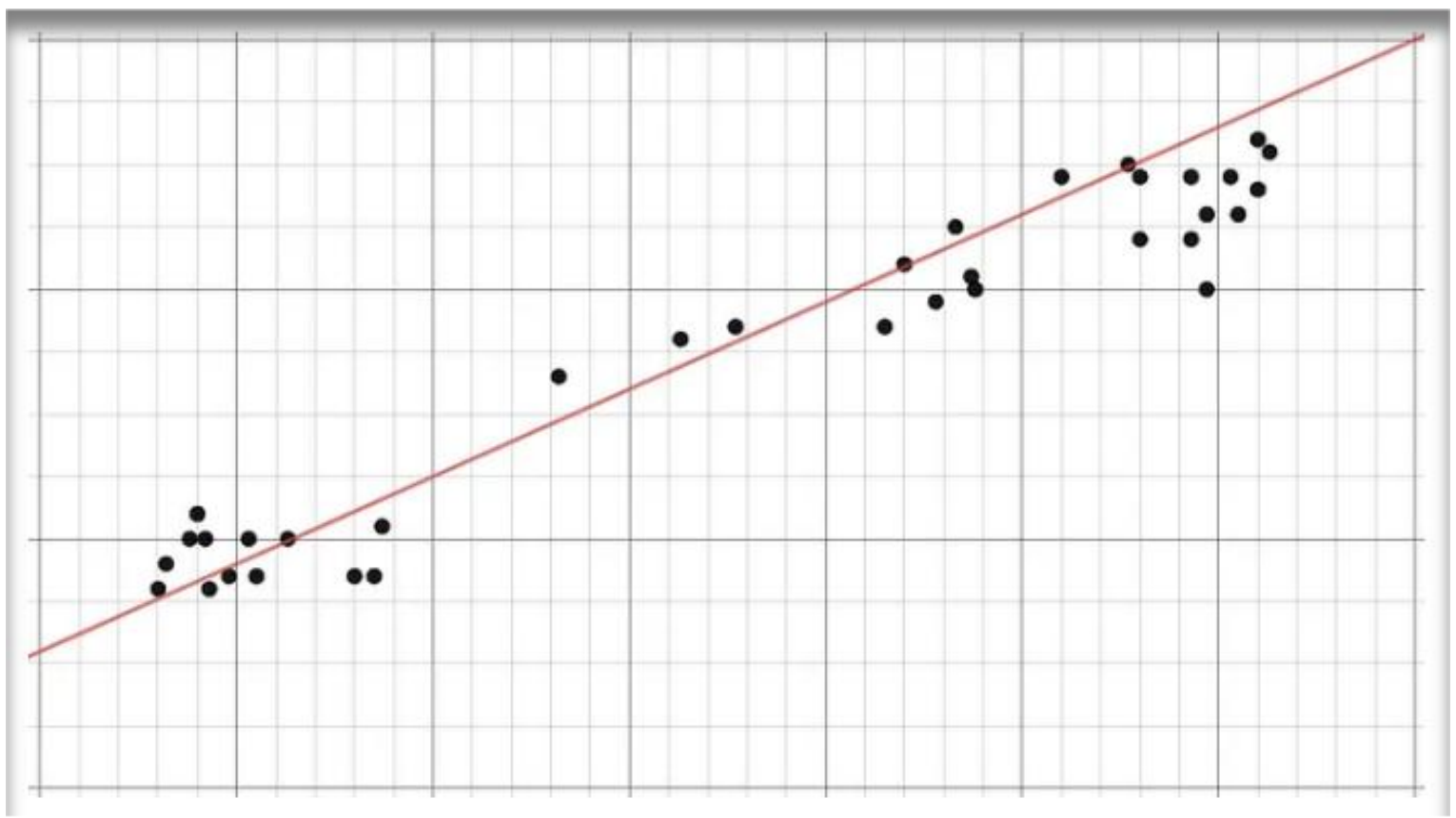

Fig 2.4 Graph showing Linear Regression [Image source – (Shin, 2022)]

## 2.9.2 Logistic Regression

Logistic regression approaches the target problem on same basic principle from linear regression and models the probability of a finite number of outcomes, often two, hence it is mostly used for binary classification assignments in which the goal is to classify cases into one of two groups. It employs a logistic function which flattens all the predicted values in the range of (0,1) to separate an instance belonging to a specific class. Logistic regression is frequently used in tasks including spam classification, churn prediction, and medical diagnosis (Rana et al., 2022). Basically, a logistic equation is made so that the outputs can only be between 0 and 1 (see below).

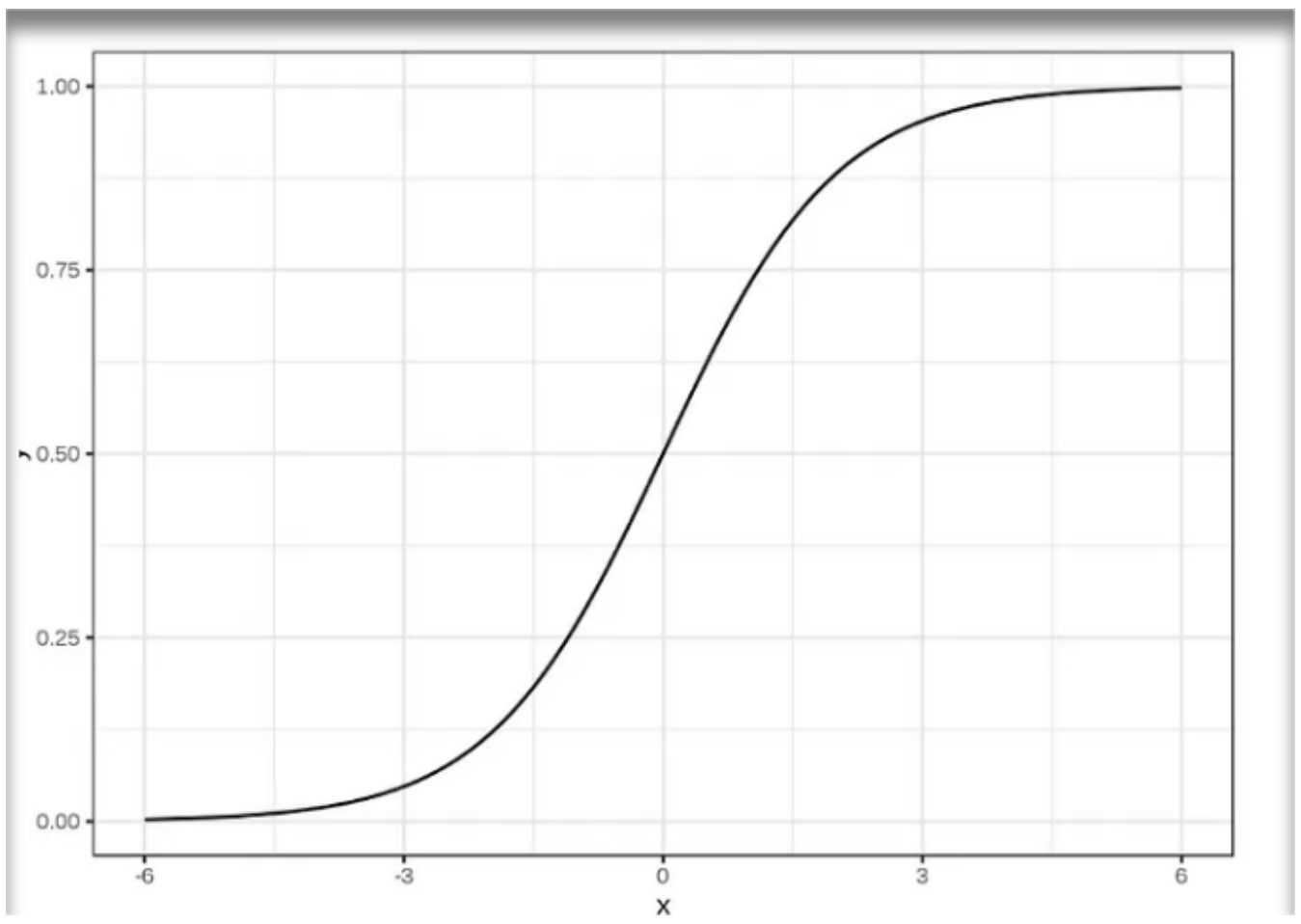

Fig 2.5 Graph showing Logistic Regression [Image source – (Shin, 2022)]

## 2.9.3 Neural Networks

Neural networks are ML techniques inspired by the operation of human brain. Similar to neurons in humans, neural networks have neurons each representing a feature from the data. These neurons pass the feature data points through the hidden layers (if present) using random weights and bias. The hidden layers assist in intense understanding of the patterns and relationships underlying the data. The network then makes the corresponding prediction and produces the result through an output later (LeCun et al., 2015)

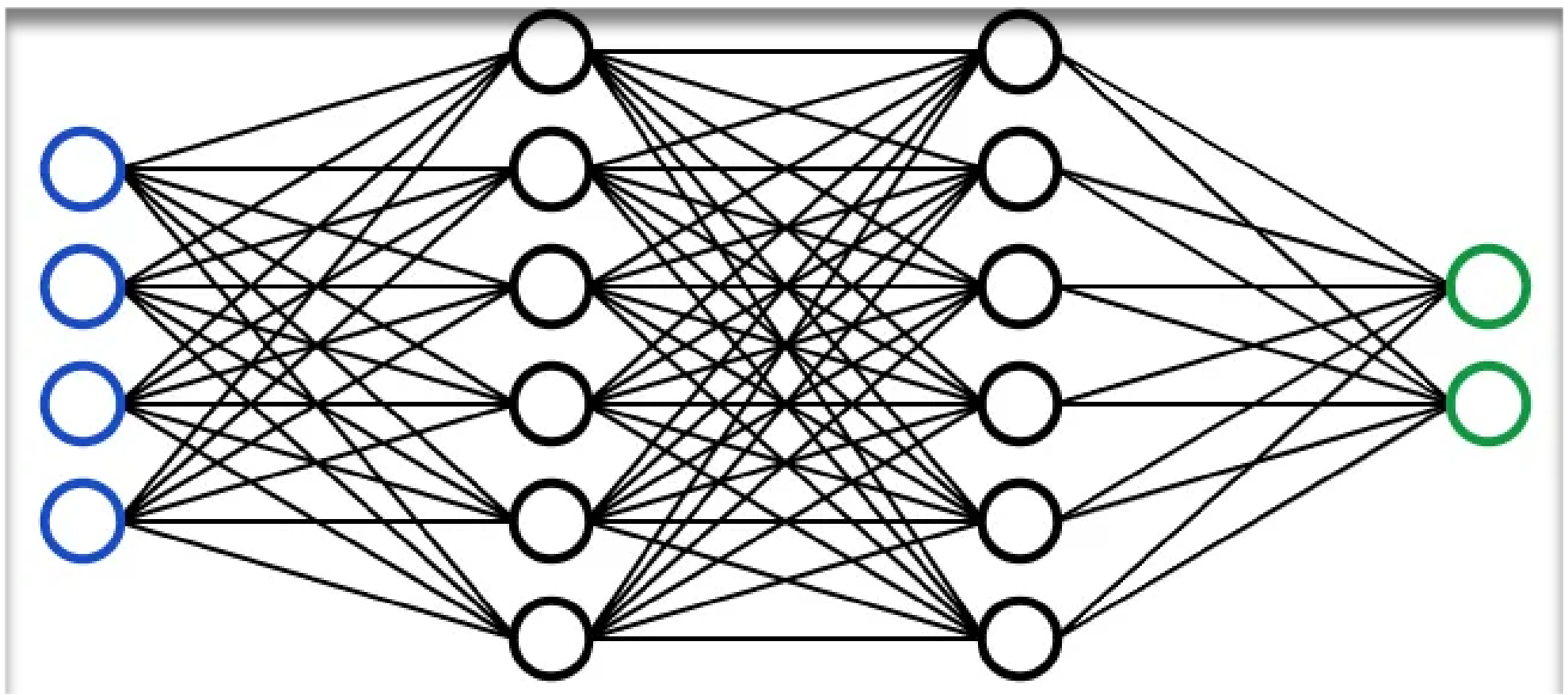

Fig 2.6 Multi-Layer Neural Network [Image source – (Shin, 2022)]

blue circles – input layer
black circles – hidden layers
green circles – output layer, output
Each node in the hidden layer has a bias and a weight from the preceding layers.

## 2.9.4 Support Vector Machine (SVM)

Often called as Support Vector Classifier (SVC), SVM is a supervised learning algorithm primarily utilised for binary classification tasks. Its primary purpose is to identify a hyperplane that optimally separates and maximises the gap between two classes. This hyperplane is determined by the support vectors, which are the data points nearest to the decision boundary (Bao et al., 2023).

Additionally, SVM can be utilised for multi-class classification and regression analysis. Image identification, text categorisation, and bioinformatics are just a few examples of the applications that make extensive use of SVM, which is famous for its capacity to manage large amounts of data.

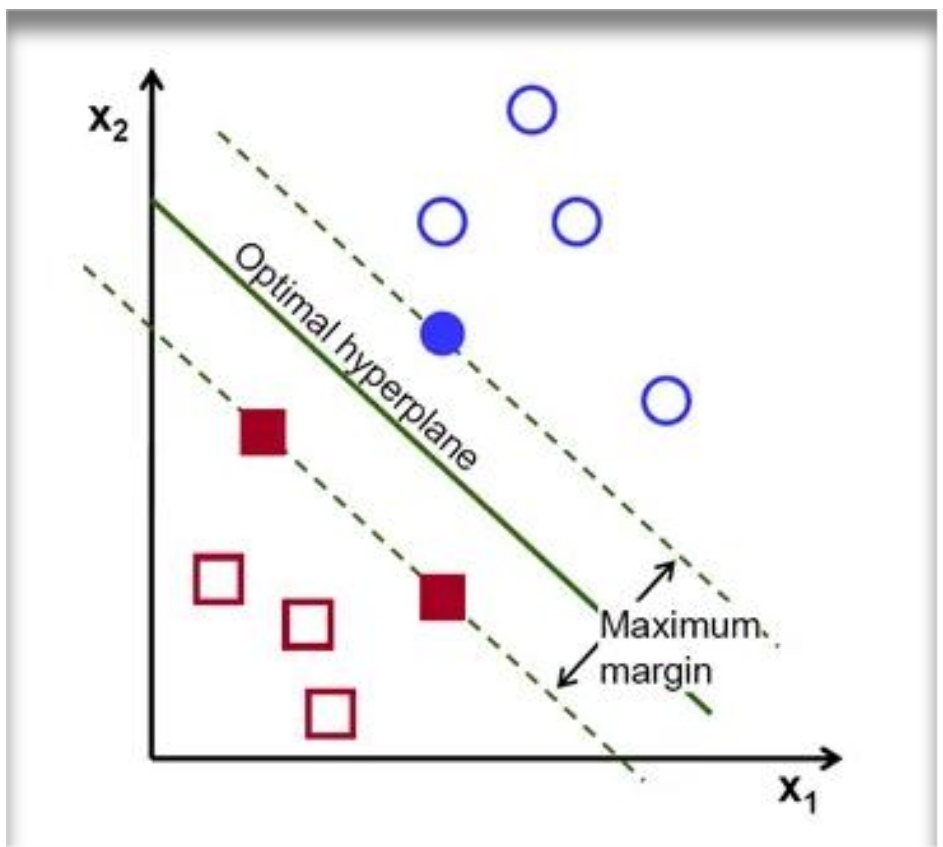

Fig 2.7 Hyperplanes in SVM [Image source – (Shin, 2022)]

From unsupervised learning, the author considered –

### 2.9.5 Clustering

The process of clustering, also known as the cluster algorithm, finds natural groupings in a dataset. It analyses the input data and looks for clusters so that the data can be separated according to the kind, nature, or form of the data. It is an unsupervised algorithm and is employed in situations when there is no class defined that needs to be predicted but the data need to be grouped. It is a technique for analysing data that is frequently used for the purpose of uncovering interesting patterns in the data, such as consumer groups that are based on their behaviour. All of the goals of cluster analysis hinge on the degree of similarity (or dissimilarity) between the many elements that are being clustered together. A clustering algorithm makes an effort to classify items in accordance with the notion of similarity that has been provided (Rodriguez-Sanchez et al., 2021)

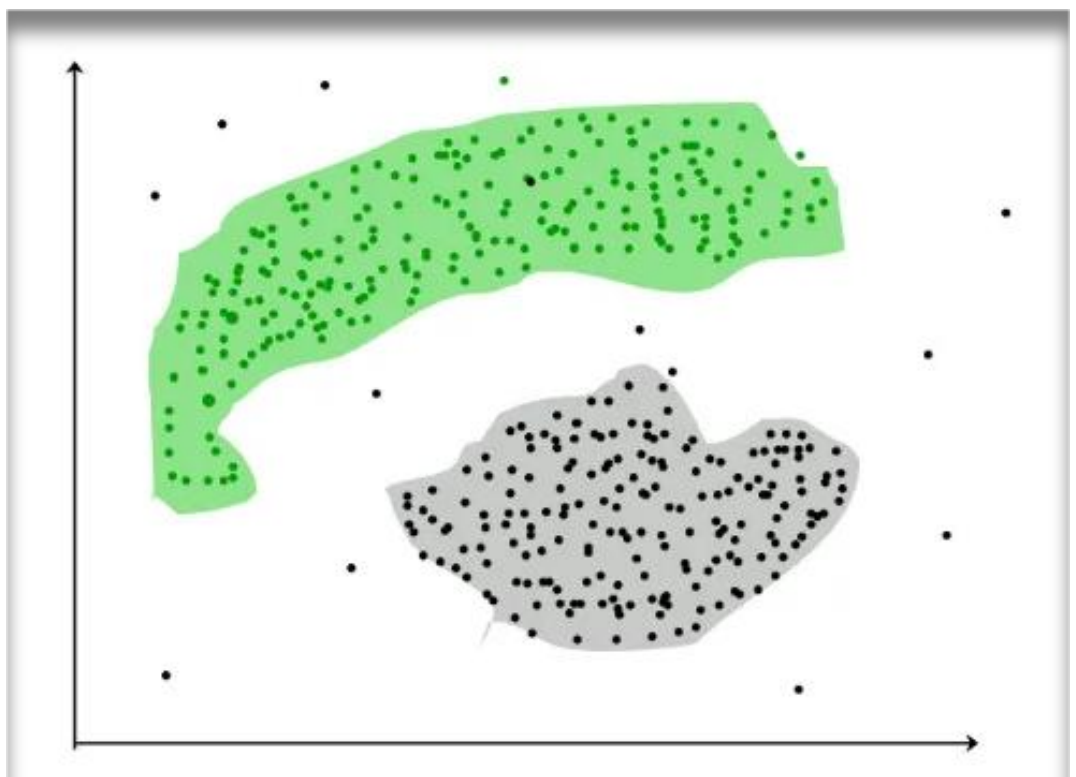

Fig 2.8 Data bifurcated into separate clusters [Image source – (Shin, 2022)]

### 2.9.6 Decision trees

The decision tree algorithm is a form of supervised learning approach that learns by analysing pairs of inputs and outputs and correspondingly translates an unobserved input to an output. It makes decisions based on a predefined set of conditions, using a tree-like architecture. The tree is composed of internal nodes, which indicate characteristics or attributes, and leaf nodes, which reflect outcomes or class labels. Each internal node divides the data based on a certain attribute in an effort to increase information gain or reduce impurity. The predictive power of the model derives from its capacity to partition the feature space into regions, hence enabling precise classification (Helmy et al., 2022).

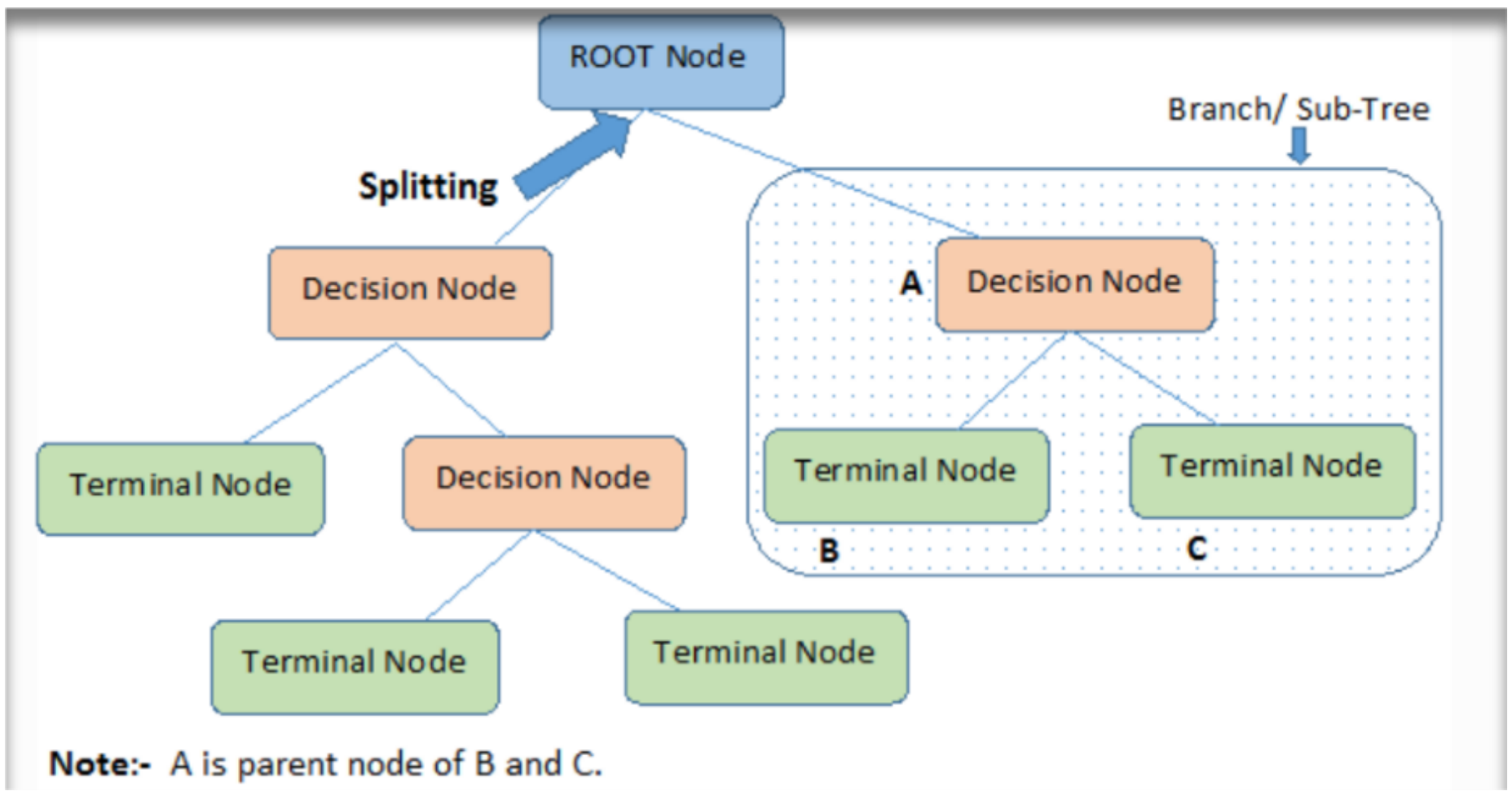

Fig 2.9 Data bifurcated into separate clusters [Image source – (Shin, 2022)]

## 2.10 Comparison of ML models

Comparative studies of ML approaches to diagnose PD.

Table – 2.4

| ML algorithm used | Type of Data | No. of test subjects | Accuracy (%) | Reference |
|---|---|---|---|---|
| SVM | Gait cycle | 64 | 83.33 | (S. Shetty and Y. S. Rao, 2016) |
| SVM with FS | Speech | 23 + 8 | 93.84 | (Karapinar Senturk, 2020) |
| Multi-class SVM with 10-fold cross-validation | MRI | 50 + 50 + 50 | 86-87 | (Singh et al., 2016) |
| ANN with FS | Speech | 23 + 8 | 91.54 | (Karapinar Senturk, 2020) |
| SVM, KNN, Ensemble AdaBoost*** | Handwriting | 37 + 38 | 81 | (Drotár et al., 2016) |
| CNN*** | Handwriting | 438 + 207 | 97.2 | (Wenzel et al., 2019) |
| CNN with 5-fold cross validation | Hand drawings | 62 + 15 | 96.50 | (Gil-Martín et al., 2019) |
| CART with FS | Speech | 23 + 8 | 90.76 | (Karapinar Senturk, 2020) |
| Ensemble RF and XGBoost | Tremor | 398 | 84 | (Xing et al., 2022) |
| RESNET50 | Hand drawings | 51 + 51 | 96.79 | (Varalakshmi et al., 2022) |
| RESNET50 + SVM (Hybrid) | Hand drawings | 51 + 51 | 98.45 | (Varalakshmi et al., 2022) |
| RESNET + MLP | Hand drawings | 51 + 51 | 98.36 | (Varalakshmi et al., 2022) |

*** are not discussed in this study but are included in the table only for comparison purposes.

## 2.11 Literature review concluded

The author observed that ML techniques such as SVM, ANNS, CNNs, Decision Trees and other deep learning algorithms are being increasingly used in PD diagnosis. By offering more precise and effective diagnostic tools, the application of ML approaches in PD diagnosis has the potential to revolutionise the clinical practice. These methods can improve early PD identification, individualised PD treatment strategies, and the disease progression tracking, which would eventually improve patient outcomes and healthcare decision-making. Critically, even though, these techniques have shown promising results in terms of accuracy and performance on specific data sets, they had some shortcomings, including small datasets and lack of standardised diagnostic criteria, among others. Furthermore, there is a need for additional validation of these algorithms by using larger and more diverse datasets to assess the generalisability and real-world clinical usefulness of ML.

Appropriate to mention here that after encompassing the benefits and shortcomings of the application of ML to the diagnosis of PD by critically reviewing the previous work on the subject matter, the author came to the logical conclusion that there is a vast opportunity for additional study especially since no ML technique has been created expressly for PD diagnosis.

# Chapter – 3 REQUIREMENT ANALYSIS

Requirement analysis is a crucial phase of software development process which primarily involves prioritising its requirements. It must be done in order to mitigate the probabilities of costly software failures. It is acknowledged as being one of the most significant decision-making processes used to examine the requirements of a project. Prioritisation provides numerous advantages, such as minimising the likelihood of misallocation of precious project resources. It also facilitates clients' and developers' comprehensions of a project's primary objectives (Khan et al., 2015).

The author in this project developed a system based on ML for the diagnosis of PD. The technology was used to analyse patient data and make precise predictions regarding the presence or absence of PD. The objective was to assist medical fraternity with early diagnosis of PD and reduce the current misdiagnosis rate. This project of developing several ML models was designed using Python language where the author used open-source libraries such as scikit-learn for ML and matplotlib for data visualisation.

Software developers and managers have suggested numerous techniques to analyse the requirements of any particular project. A careful attention and time were allotted by the author to a few of such methods such as analytic network process (ANP), scrum, agile, waterfall, analytic hierarchy process (AHP), hierarchy AHP, MoSCoW, spanning tree matrix, bubble sort, binary search tree and priority groups. After careful consideration and study of these methods, the author decided to use MoSCoW which has a different approach towards the requirement analysis.

| Evaluation Criteria | Simple Ranking | MoScoW | 100 dollar | AHP |
|---|---|---|---|---|
| Ratio Scale Information | | | Yes | Yes |
| High Confidence from User | Yes | Yes | Yes | |
| Consistent | Yes | Yes | Yes | Yes |
| Low difficulty | Yes | Yes | Yes | |
| Low effort | Yes | Yes | Yes | |
| Able to handle large number of alternatives | | Yes | | |

Table – 3.1 Comparison of multiple techniques [Source – ( Khan et al., 2015)]

## *3.1 Museum of Soviet Calculators on the Web*

***MoSCoW*** is an acronym that stands for four hierarchical priority groupings. Each requirement in a group is assigned the same priority. This method categorises requirements into four groups based on their priority: MUST-have, SHOULD-have, COULD-have, and WON'T-have (Black, 2020). Each criterion is assigned to one of the groups depending on its relative importance.

MUST-*have* – this group of requirements have to be present in the software prior to its release. These requirements are non-negotiable; failing to meet them will result in the failure of the entire project.

SHOULD-*have* – the product/software will benefit from the implementation of these requirements. Optional features that would be a plus if they could be included.

COULD-*have* – fulfilled requirement from this group are beneficial to the project/software. Features that would be beneficial to have if at all possible but slightly less advantageous than the SHOULD-have.

WON'T-*have* – these needs cannot be implemented in the current iteration since they are low priority. This is sometimes referred to as a "wish list." These criteria are not inconsequential, but they may be implemented at a later time.

(Black, 2020; Overby, 2021)

Therefore, based on MoSCoW the author categorised and prioritised the *specific requirements* of this project into four groups stated below –

***MUST-have***

The stages of getting access to the dataset, choosing the right ML algorithms, putting data preparation and preprocessing techniques into practice, including data cleaning and feature selection, and assessing the performance of the developed model using the right metrics were the non-negotiables (must-haves) for this project.

Table – 3.2

| **Requirement** | **Priority** |
|---|---|
| Data Collection | *High* |
| Data Preprocessing | *High* |
| Feature Selection | *Medium* |
| Model Selection and Development | *High* |
| Model Evaluation | *High* |
| Model Optimization | *High* |
| Model Validation | *High* |

***SHOULD-have***

In addition to the above-mentioned non-negotiable elements, the author utilised various techniques to improve the project's overall performance by comparing various model evaluation strategies, researching multiple ML algorithms and finally used that was found to be most suitable for this project, and applied various feature extraction techniques as well.

Table – 3.3

| Requirement | Priority |
|---|---|
| Comparison to other models | *High* |
| Research multiple ML methods | *Medium* |
| Feature Extraction Techniques | *Medium* |

***COULD-have***

The author could have tested the models on multiple datasets from different sources to improve the accuracy and reliability of the diagnostic model. Further, attempts were made to developed ensemble models by combining different ML methods to enhance the model's performance. Incorporation of complex deep learning or neural network models were also considered.

***WON'T-have***

Commercial datasets or tools that were unavailable or impractical for the study. In addition, the system does not have a user interface.

## 3.2 Hardware

The author used his personal computer having an Intel Core i7 processor with 500 GBs of SSD and 16 GBs of ram which was sufficient for carrying out coding part of the experiments. Although, running large number of epochs for some experiments did turn the machine slow but it was manageable.

# Chapter – 4 PLES

***(PROFESSIONAL, LEGAL, ETHICAL AND SOCIAL ISSUES)***

## 4.1 Professional Issues

The author wrote and tested the code while adhering to British Computing Society (BCS) code of conduct. Code was written to a high standard and commented throughout for clarity. There is adequate documentation presented. All third-party software, libraries, and other code fragments were utilised only if they were open to use. Any citations or outside data was properly referenced.

## 4.2 Legal Issues

Exploring the PD diagnosis in the context of data science required the use of sensitive patient data, so the author followed all the necessary and proper data privacy and security measures in accordance with General Data Protection Regulation (GDPR) UK.

## 4.3 Ethical Issues

When working with data received from real subjects, it is imperative to consider ethical considerations such as informed consent, data privacy, and data security (Lamba et al., 2022).

In particular, the author deliberated that –

- the research was conducted with integrity and transparency.
- the data was managed carefully, respecting the rights of test subjects, supervisors, and Heriot Watt University as well as their right to privacy.
- the participants' data was kept confidential.
- the information was securely kept and guarded against unauthorised access or disclosure.

Last, the author collected the data with care and managed it in accordance with the data privacy and security standards outlined by GDPR.

## 4.4 Social Issues

As there were no direct communication with any patient or test subjects so the project didn't come across any social issues.

# Chapter – 5 IMPLEMENTATION

In this section, the author has described the implementation and the practical execution of the solutions proposed to achieve the objectives of the study, outlined in the introduction chapters of this report and has also provided a detailed account of the implemented ideas. To begin, the author has given an overview of the methodology and techniques that were utilised during the implementation phase. As this research does not involve any physical tests, the author has accordingly made use of the necessary software which have been described along with their configurations and specifications. Additionally, the experimental setup and any modifications made to existing methodologies have also been outlined. Throughout the implementation process, the author has elaborated the challenges encountered and the various strategies and problem-solving techniques used to overcome them. The author believed that by starting the implementation chapter with a clear outline of the methodology and a step-by-step explanation of the practical steps taken, readers will gain a comprehensive understanding of how the research was brought to life.

Therefore, in this section, the author provides an overview of the experiment and different ML approaches which were used to analyse the dataset to explore PD. The author has broadly followed the steps mentioned below to develop and apply multiple ML techniques for this study –

- Data Collection
- Data Preprocessing
- Feature Selection
- Model Selection and Development
- Model Evaluation
- Model Optimisation

## 5.1 Platforms used for experimentation

The author has successfully executed the proposed solution and conducted the research using various tools and platforms.

Coding language opted – Python

Data analysis – *MS excel* was used for a preliminary examination of the data leveraging its powerful analytical capabilities and data manipulation features which enabled the author to gain valuable insights and perform the necessary preprocessing steps.

Data storage – although the data was available locally on the machine, but the author decided to store a copy of the data in their private *GitHub repository* because the university GitLab was unavailable around the time, and it allowed for seamless integration and version control.

https://github.com/farhaanqazi/F21MP

Model training and testing – *Google Colab* was used to train, test and evaluate all the ML models with the help of different python libraries. Colab continuously saved the work done or any changes made to the google cloud, thus protecting from any loss of work. It also provided the author an easy access to the code, after proper authorisation.

https://colab.research.google.com/drive/1rrxFW2hT65NId1ape838EWG_zUk2TAOH

## 5.2 Collection of Data

The author has used a dataset which was collected at Leeds Teaching Hospitals NHS Trust for this research. The dataset was received by the author from the university through approved channels.

## *5.2.1 Dataset explained*

The dataset has information from 58 people comprising of confirmed PD patients and healthy controls. The data was created by asking the test subjects to trace and draw spiral pentagons and cubes on a digitising pressure-sensitive Wacom tablet (Wacom Technology Corporation) of size (20.3*32.5) cm. Wherein each subject was asked to draw both the figures once by each hand i.e., by their dominant and non-dominant hand, thereby constituting 4 rows of data from each test subject. They were instructed that the figures should be drawn as quickly and accurately as possible (Alissa et al., 2022).

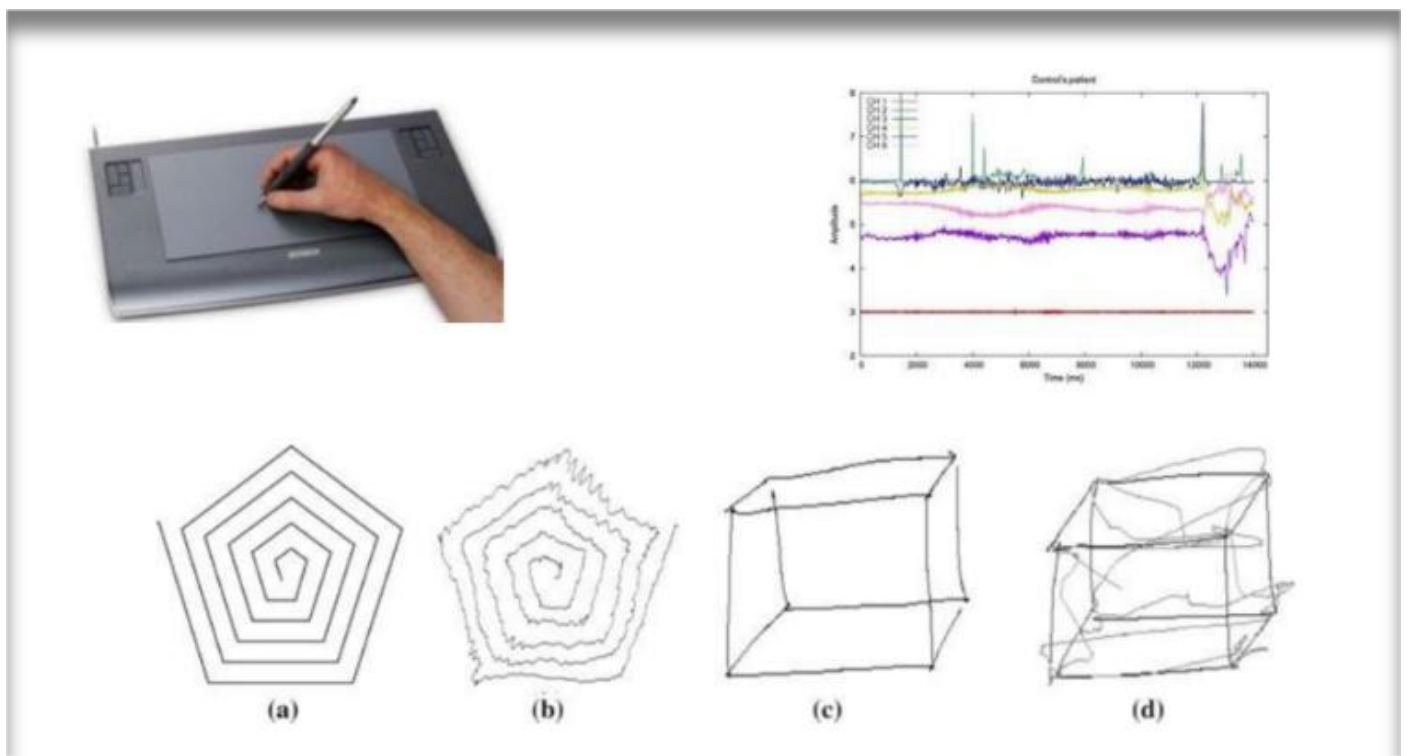

Fig 5.1 Data Gathering Method

The data has 19 columns  and 325 rows. A few important features that were used by the author while writing the code for the analytical algorithms were “Dominant”, “Attempts”, “Duration”, “Time”, “AreaError”, “LeaveSurface”, and “TimeContact”.

As a suitable and ethically approved dataset was already available, so the author did not need to gather more data and used the already collected and verified data for the study. Moreover, it is considered suitable for this type of student-based and completely non-funded study.

## 5.2.2 Exploratory Data Analysis

During the initial overview of the dataset, the author established that the target column 'PC', which defines the PD patients and healthy controls as 1 and 0 respectively, have 55 samples from confirmed PD patients and only 25 samples from healthy controls. As the samples in the two classes are not roughly equal, the dataset can be referred to as unbalanced. For unbalanced dataset used, the performance of ML models can be inclined towards the confirmed PD patients. The model may have a higher accuracy in predicting the confirmed PD patients (majority class) but may perform poorly in predicting the healthy controls (minority class).

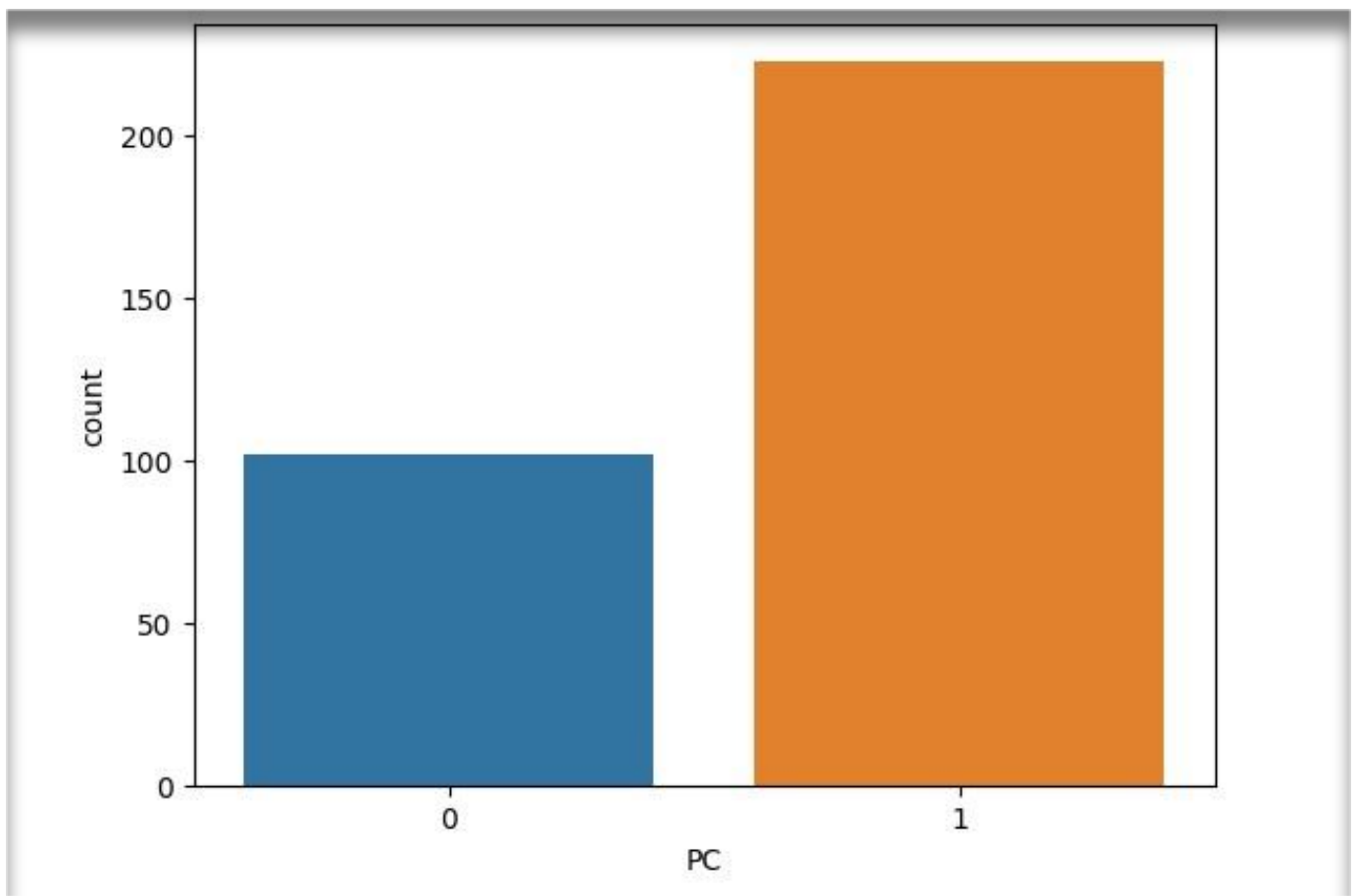

Fig 5.2 Target Column Class Count from Dataset

The author also endeavoured to visualise the data features by plotting all the columns.

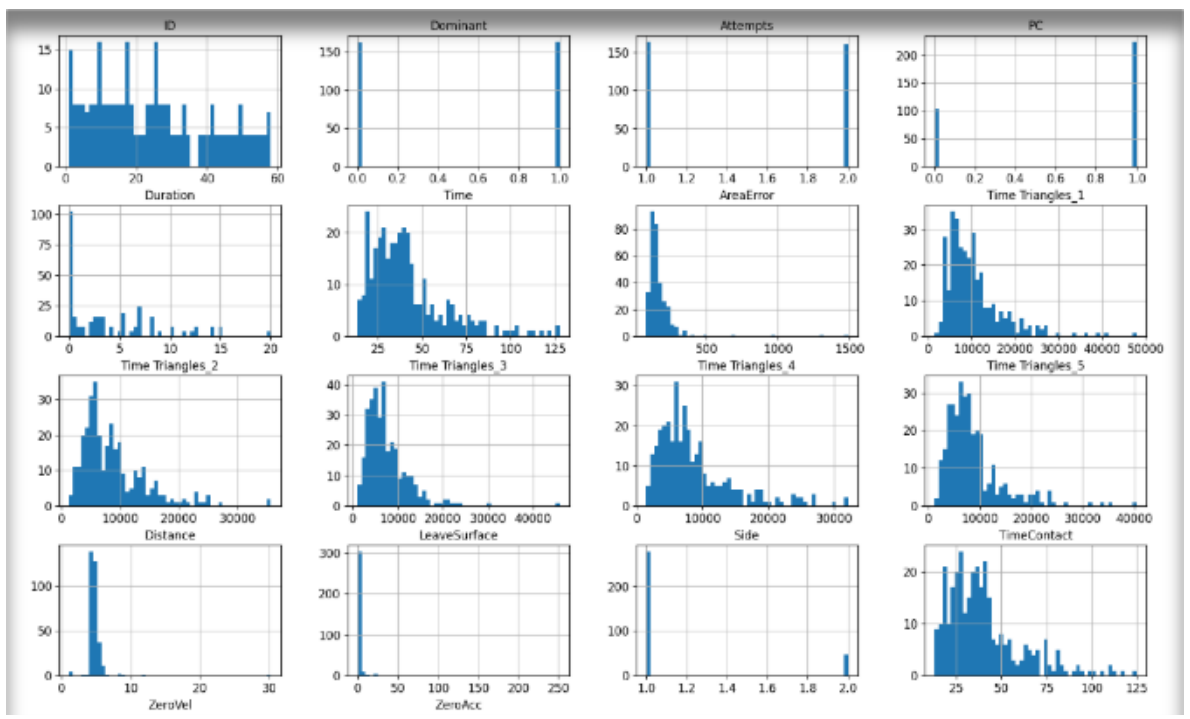

Fig 5.3 Histograms of dataset features

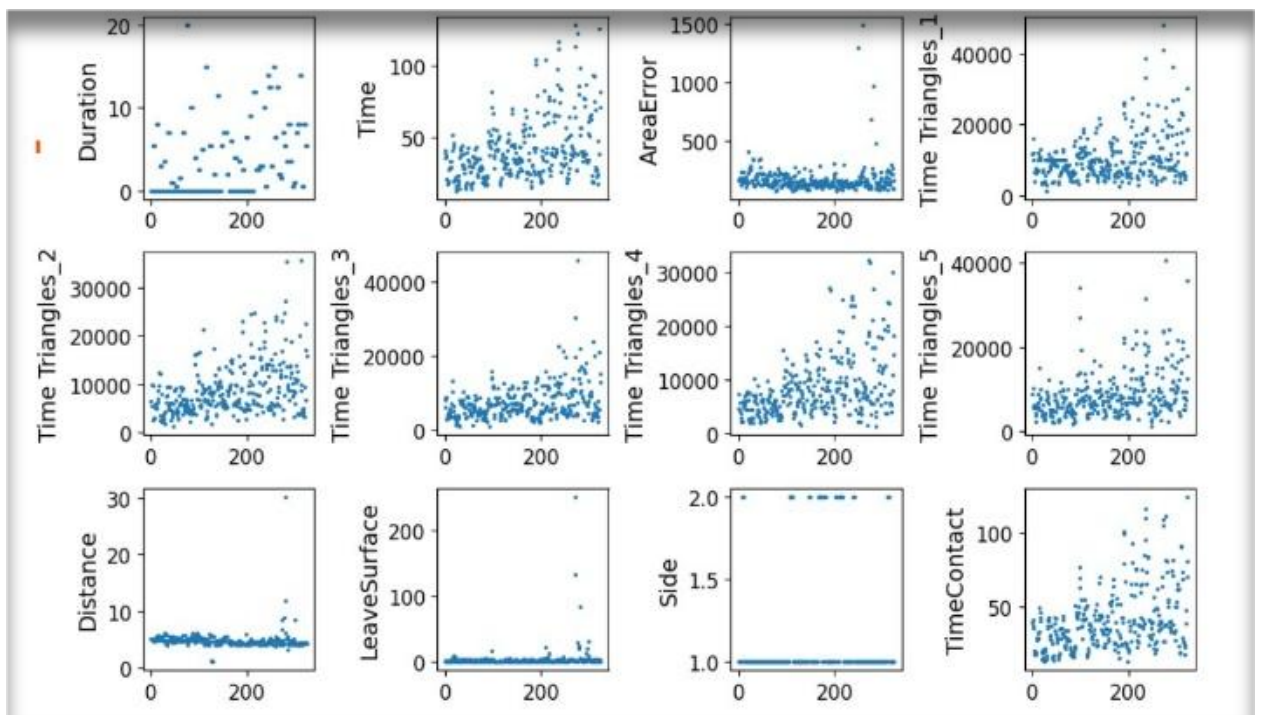

Fig 5.4 Scatter Plots of dataset features

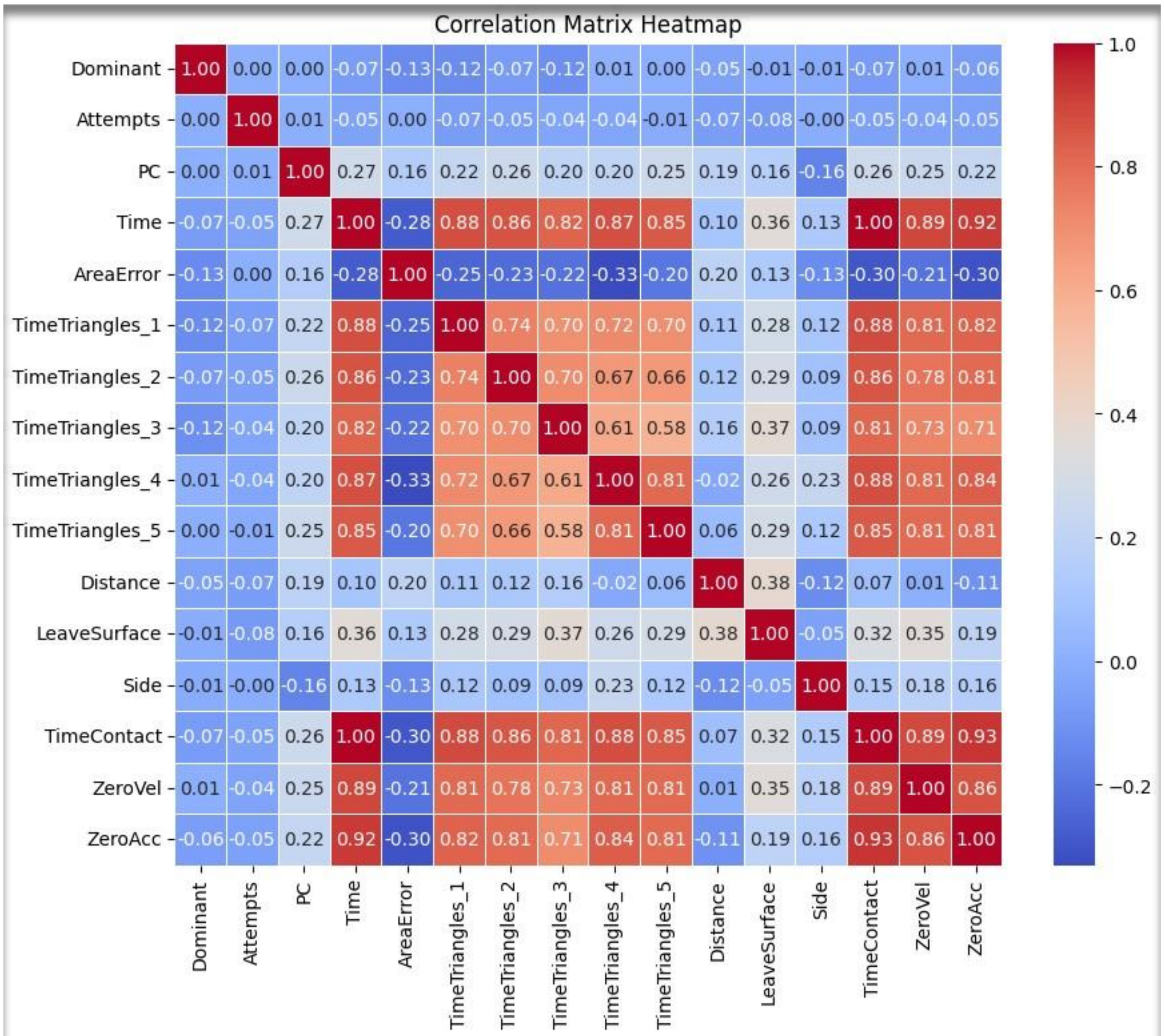

Fig 5.5 Correlation Matrix Heatmap

## 5.3 Data Preprocessing

Subsequent to the successful insertion of the dataset to the algorithm using the GitHub repository link to the data, the author began the preprocess procedure. The purpose of preprocessing is to prepare the data aiding an improved use of the gathered data for the training of an ML model (Paleyes et al., 2022). The data was examined for cleanliness and missing values, but the dataset was found to be clean without any missing values. At this stage, author also examined for any outliers and unnecessary data points from the data. All the columns in the dataset were explored for any outliers by plotting a boxplot and analysing it visually. The author also standardised the data to assure that each feature was given the same weight.

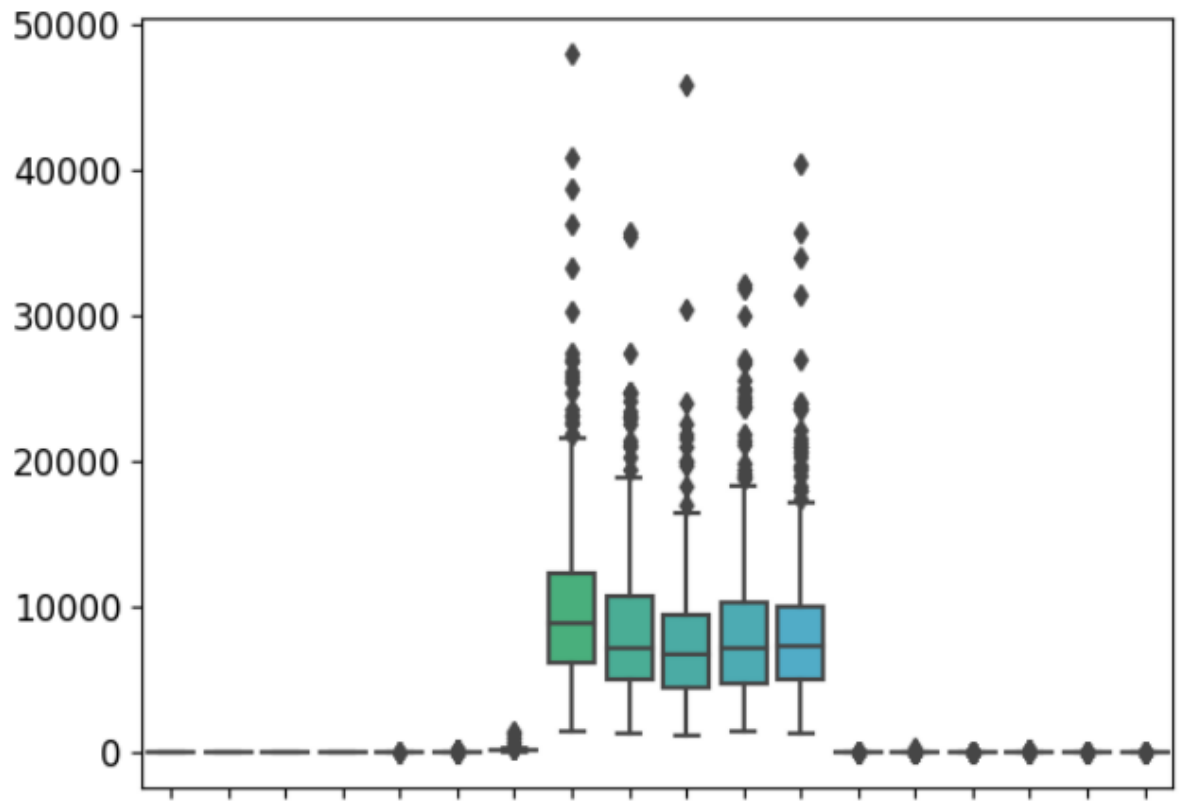

Fig 5.6 Box plot for outliers

As outliers can significantly impact the accuracy and reliability of statistical analyses, the *z-score* method was employed by the author as a robust technique to identify and remove outliers from the dataset (Lee and Styczynski, 2018).

```
 # Calculate the z-scores for the column
   z_scores = np.abs((PDdata[column] - PDdata[column].mean()) /
PDdata[column].std())

# Find the indices of outliers in the column
    outliers = PDdata[column][z_scores > threshold]
    outlier_indices.extend(outliers.index.tolist())
```

By calculating the z-scores of each data point, the author was able to detect extreme values in each column that deviated substantially from the mean. These outliers were subsequently eliminated from the dataset, ensuring a more representative sample for further analysis. To address the missing values created by outlier removal, the *KNN imputation* method was engaged. KNN imputer utilises the information from neighbouring data points to estimate and replace the missing values, thus maintaining the integrity of the dataset and minimizing bias in subsequent analyses (Lee and Styczynski, 2018).

```
# Apply KNN imputation to handle missing values
knn_imputer = KNNImputer()
PDdata_imputed_data = pd.df(pdknn_imputer.fit_transform(PDdata),
columns=PDdata.columns)
```

```
# Print the replaced data
print("Replaced Data:")
print(PDdata_imputed_data)
```

Importantly, the author also discovered that the column "ID" is just an identification number of the test subjects, hence cannot be a possible feature for making predictions.

```
print (df['ID'])

0      10
1      10
2      10
3      10
4       1
       ..
320     8
321     9
322     9
323     9
324     9
Name: ID, Length: 325, dtype: int64
```

Fig 5.7 Description of the column 'ID'

Additionally, the column "Duration" has a strong association with the target column "PC" and causes overfitting across all the algorithms. Therefore, the columns had to be dropped prior to employing ML classification algorithms.

```
combined_columns = pd.concat
print(combined_columns)

     Duration  PC
0         0.0   0
1         0.0   0
2         0.0   0
3         0.0   0
4         0.0   0
..        ...  ..
320       1.5   1
321       7.0   1
322       7.0   1
323       7.0   1
324       7.0   1

[325 rows x 2 columns]
```

Fig 5.8 Association of columns 'Duration' and 'PC'

```
#these columns caused overfitting
PDdata.drop(["Duration"], axis=1, inplace=True)
PDdata.drop(["ID"], axis=1, inplace=True)
```

## 5.4 Feature Selection

In data mining, feature selection is a common practise used to clean up datasets by omitting irrelevant information. This method enhances data readability, the efficiency with which learning algorithms are trained, and the accuracy with which they make predictions. Consequently, at this stage, all the relevant features that were realised as most relevant were selected from the preprocessed data. Because the author has used more than one ML method, features were selected based on the algorithms that was being trained. Feature selection is discussed further in the model development section.

## 5.5 Model Development

The complexity of the data is often the deciding element in selecting an appropriate ML model to be applied in a given scenario. While deep learning and reinforcement learning have gained popularity in academic circles, simpler models are still frequently employed in industry (Paleyes et al., 2022). In the context of model selection, as the primary objective of this dissertation was to explore the PD diagnosis using ML techniques, so a single model couldn't have sufficed the purpose of it. So, evidently multiple algorithms were studied and applied to the preprocessed data after choosing the appropriate features.

For the development of a predictive model, dataset needs to be split into training and test samples. Training data is fed to the selected model which enables it to identify and learn the patterns in the data. The model is then tested on the test data, still unidentified to the newly developed model (Paleyes et al., 2022).

Accordingly, the author split the dataset into training and testing samples, comprising 80% and 20% of the data respectively. The same train and test samples were used for all the model developments.

```
# Split the data into training and testing sets
X_train, X_test, y_train, y_test = train_test_split(X, y,
test_size=0.2)
```

Here, the author has comprised a comprehensive explanation of each ML algorithm that was used to evaluate its efficiency on the acquired dataset. The explanation begins with a condensed description of the model development as well as its operation, and then moves on to examine its applicability to the dataset. Further, the results obtained for each algorithm with an explanation are also provided in subsequent chapters.

### *5.5.1 Decision Trees*

The decision tree algorithm is a form of supervised learning approach that learns by analysing pairs of inputs and outputs and correspondingly translates an unobserved input to an output. It makes decisions based on a predefined set of conditions, using a tree-like architecture. The tree is composed of internal nodes, which indicate characteristics or attributes, and leaf nodes, which reflect outcomes or class labels. Each internal node divides the data based on a certain attribute in an effort to increase information gain or reduce impurity. The predictive power of the model derives from its capacity to partition the feature space into regions, hence enabling precise classification (Exarchos et al., 2012). After splitting the dataset into train and test sets, and feeding the training data to decision tree algorithm, it recursively partitions the feature space by choosing the most informative attributes to create decision rules.

*Development*

Accordingly, the author began the model development by feeding the already split training data to the algorithm. By creating an instance of the decision tree classifier in scikit-learn library of Python, the model is thereby trained on the test samples of the data.

```
# Train a decision tree classifier on the training data
model = DecisionTreeClassifier(max_depth=5)
```

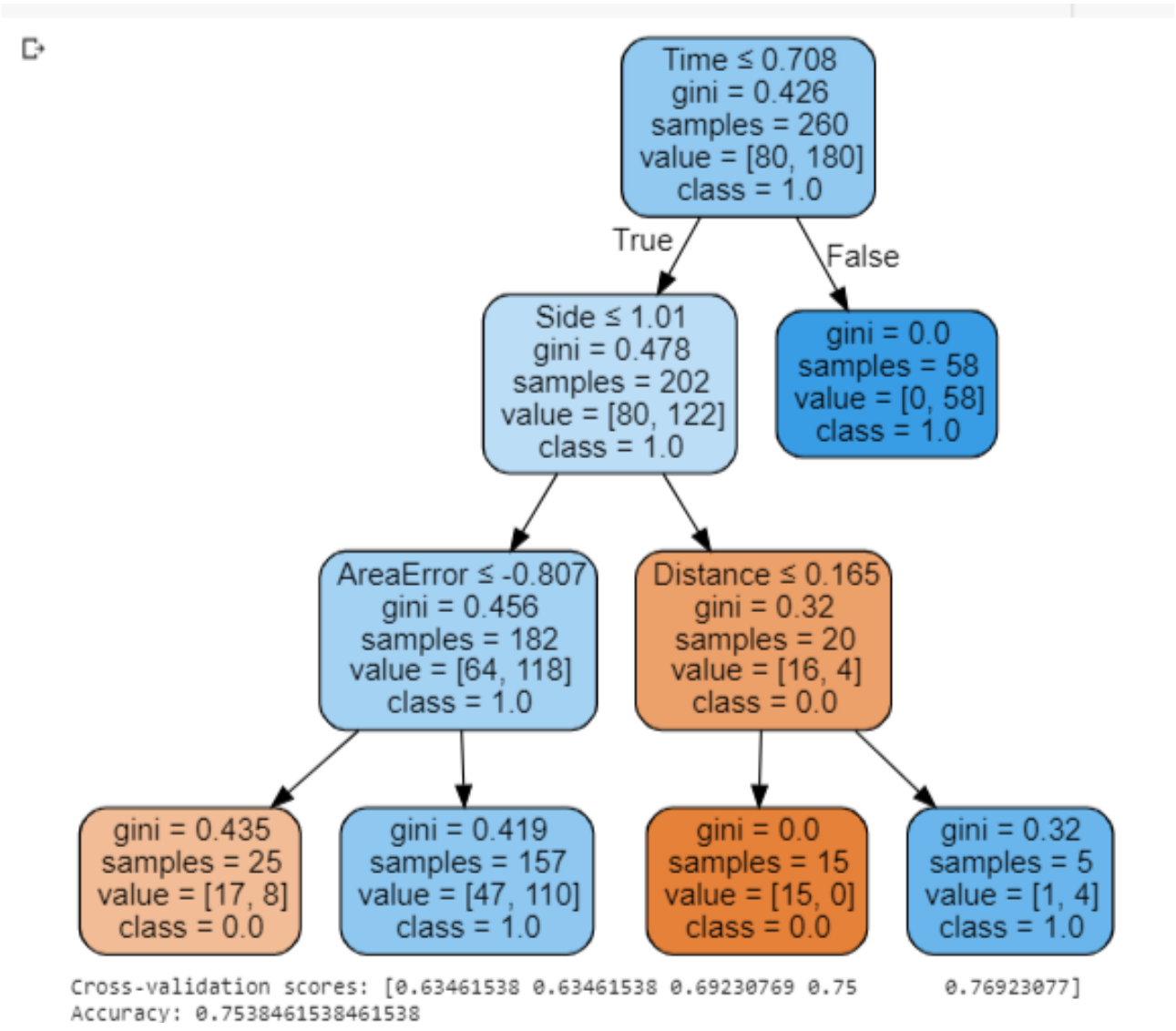

Fig 5.9 Decision tree produced from the algorithm

Above tree was produced after running the decision tree algorithm with a *max_depth* of 5. It consists of 6 nodes representing the features - Time, Side, AreaError, and Distances and Gini. In decision trees, all the features are given a number (Gini index) based on their relative usefulness on the predictive capability of the model. Gini index is calculated from the sum of the squared probabilities of each class and is used to determine the best feature for splitting the data. The feature with the lowest Gini index is chosen as it leads to the strongest child nodes (Exarchos et al., 2012). The values of the features and their gini indices determine the path the data point takes through the tree and the final nodes provide the predicted label (0 or 1) for each data point.

```
# Train the model
model.fit(X_train, y_train)
```

And so, the tree was trained on a subset of the dataset with *260* samples.

### 5.5.2 Multilayer Perceptron

In ML and pattern recognition, the Multilayer Perceptron (MLP) classifier is a popular feedforward ANN architecture for supervised learning tasks. When the data flows only in one direction starting from input layer making its way towards the output layer where outcomes are produced, it is called a feedforward neural network. There are no feedback loops or recurrent connections and there may or may not be hidden layers present between the input and output layers (Pal and Mitra, 1992).

In MLP, data is fed to the model through an input layer which consists of neurons that represent the features of the input data. An MLP can have hidden/s layers which consist of numerous neurons that process the input data using learned weights and biases, allowing the model to capture complex patterns in the data. Finally, an output layer produces the final predictions of the classifier (Grippo et al., 2016).

## *Development*

After splitting the dataset into train and test samples, the author scaled the data using the *StandardScaler()* library of python. The standardisation step was essential because it centred the data around zero mean and scaled it to unit variance, hence limiting the effect of variable scales on the learning process of the model. Further, using scikit-learn library, an instance of the MLP classifier was initialised consisting of two hidden layers with *relu* function. The model initially had 250 and 200 neurons in its hidden layers. The *adam* optimiser was employed for training of the model.

```
# Create an MLP classifier
model = MLPClassifier(hidden_layer_sizes=(250, 200), activation='relu',
solver='adam', random_state=42, max_iter=200)
```

ReLU, which stands for Rectified Linear Unit, is an activation function which introduces nonlinearity to the model without triggering the vanishing gradient problem. ReLU activation enables the model to learn faster as compared to sigmoid and tanh functions (Eckle and Schmidt-Hieber, 2019). By the introduction of non-linearity, the MLP model was hence enabled to understand more complex relationships in the data.

Adaptive Moment Estimation (Adam) is an optimiser that determines the learning rate while an MLP classifier is being trained. Here, it adjusted the learning rates of each parameter based on their historical gradients, making it suitable for the small dataset on hand. Adam's customisable learning rate allows it to converge faster and more effectively than conventional stochastic gradient descent (Eckle and Schmidt-Hieber, 2019).

### *5.5.3 k-Nearest Neighbour*

The *k*-Nearest Neighbours (KNN) is a non-parametric algorithm which is characterised by not making assumptions about the data distribution. Instead, to generate predictions on new and unseen data, it memorises the entire training dataset. It makes predictions based on the majority class of its *k* nearest neighbours in the feature space. The parameter *k* represents the number of neighbours considered for each prediction (Cai et al., 2018).

### *Development*

With the python libraries, the author developed the model after scaling the features using the *StandardScaler().*

```
# Create a KNN classifier
model = KNeighborsClassifier(n_neighbors(k)=11)  #Specify the number of neighbours
```

In KNN model development, the author took the cognisance that the performance of the model largely depends on the number of neighbours taken into consideration while training the model. (Bellino et al., 2019) says that a smaller *k* value makes the algorithm more susceptible to noise, whereas a bigger *k* value may smooth out the decision boundaries and be less impacted by noise, but less discriminative. Consequently, the author took great care in determining the number of neighbours (k) when constructing the model and had to tune it numerous times to achieve promising results. In this case, a *k* value of 11 was opted for the model. It is discussed in detail under hyperparameter tuning.

## 5.5.4 Logistic Regression

In the context of PD diagnosis, from the authors perspective logistic regression was an important ML algorithm to be explored, given its classic binary categorisation functions.

Logistic Regression is a supervised learning ML technique that is well suited to classification tasks. Its primary objective is to discover the best logistic function capable of mapping input data to a binary output, indicating the presence or absence of a specific condition or occurrence. At the core of logistic regression is the sigmoid function (activation function), commonly called the logistic function which compresses any real-valued number into the interval [0, 1]. The use of an activation function is essential in neural networks in because it allows the model to interpret the output as a probability, by instituting non-linearity into the model, expressing the likelihood of a data item belonging to a specific class (Liu et al., 2020).

When applied to a binary classification with input features (X_train) and corresponding binary classes (y_train) as the training data, logistic regression optimises a set of weights and the bias to fit the training data in a way that minimises the logistic loss function. This loss function produced the error (loss) from the predicted labels and the actual ones. Once the training is complete, the model can use these learned coefficients to make predictions on new, unseen data (Liu et al., 2020).

## *Development*

After careful study and consideration, the author chose RFE for feature extraction for this particular algorithm.

```
# Feature selection using RFE with Logistic Regression estimator
estimator = LogisticRegression()
n_features_to_select = 5
rfe = RFE(estimator=estimator
X_new = rfe.fit_transform(X, y)
```

In the above code, the author has employed an effective feature selection technique, Recursive Feature Elimination (RFE), in conjunction with the Logistic Regression estimator. The primary goal was to discover the most important features from the dataset that are essential for predicting the target variable ('PC'). RFE is a well-established method for feature selection that recursively eliminates the least important characteristics from a dataset. RFE determines the top features that contribute the most to the performance of the model by iteratively executing this procedure (Lin et al., 2022).

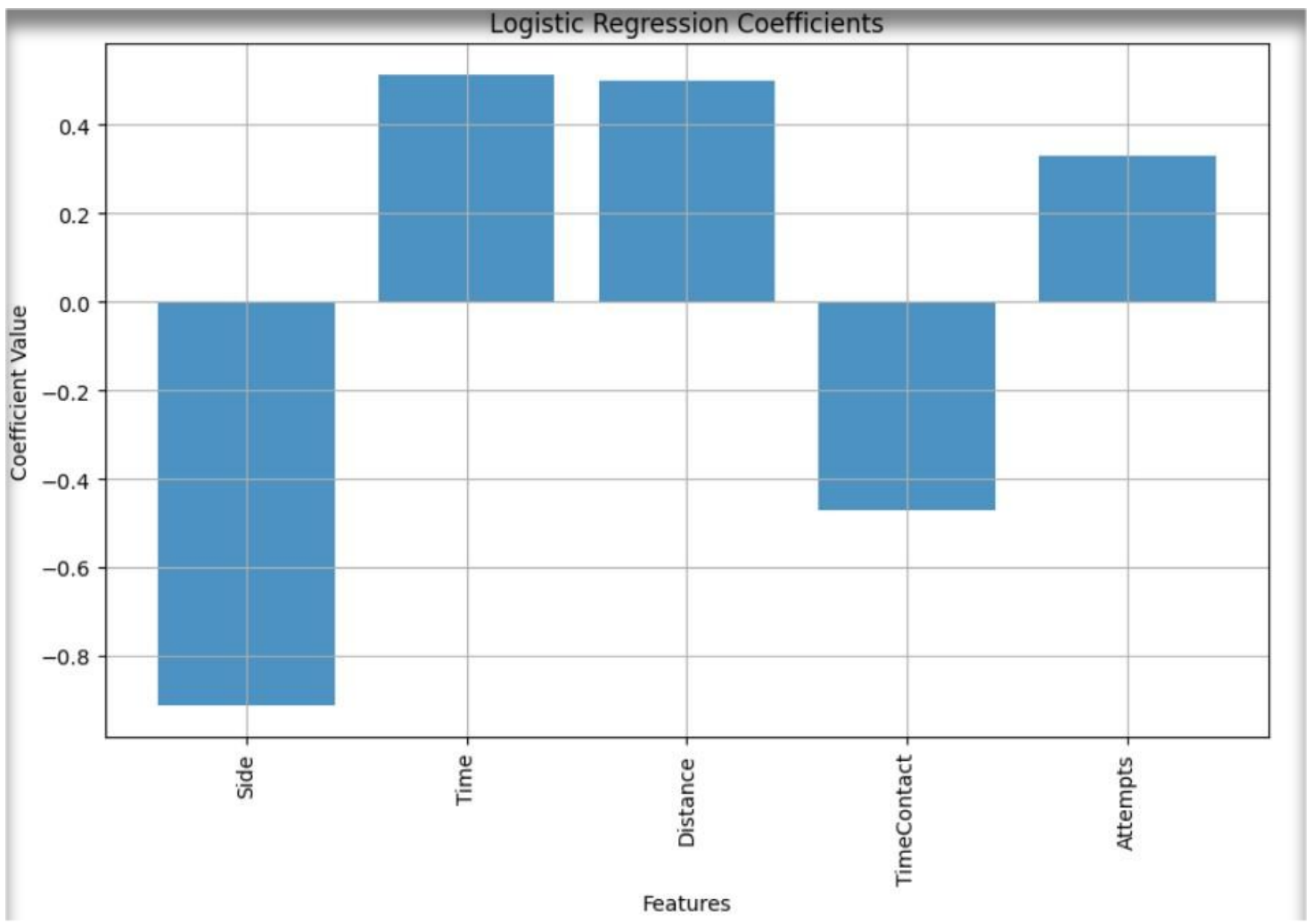

Fig 5.10 Logistic Regression Coefficients

The number of retained features, designated by 'n features to select,' was set to 5, indicating the author's desire to find the five most influential features for the classification task. This method enabled the author to optimise the performance of the model by lowering the dimensionality of the dataset while preserving the most informative characteristics.

Afterwards, the fitting of the model was initiated by creating an instance of Logistic Regression algorithm from the scikit-learn library of python.

```
# Model training
model = LogisticRegression()
model.fit(X_train, y_train)
```

Further, the model was fit on the training samples from the selected features where the model learns from the input data represented by X_train and corresponding target labels denoted by y_train. In the background, the 'fit' method automatically applies the logistic (sigmoid) function to calculate the probabilities for the binary classification task. It then optimises the model parameters to minimise the cross-entropy loss between these calculated probabilities and the actual labels.

### *5.5.5 Support Vector Mechanism*

Often called as Support Vector Classifier (SVC), SVM is a supervised learning algorithm primarily utilised for binary classification tasks. Its primary purpose is to identify a hyperplane that optimally separates and maximises the gap between two classes. This hyperplane is determined by the support vectors, which are the data points nearest to the decision boundary.

#### *Development*

After splitting the dataset into train and test samples, the author scaled the data using the *StandardScaler()* library of python similar to it was used during the development of MLP model. Further, for extracting most suitable features from the data, after careful study and consideration, the author selected Principal Component Analysis (PCA).

```
# Perform PCA for feature extraction
pd_pca = pdDataPCA(n_components=10)  # number of components
X_train_ pd_pca = pd_pca.fit_transform(X_train_scaled)
X_test_ pd_pca = pd_pca.transform(X_test_scaled)
```

Reduction of the original dataset into a smaller more readable and efficient subset of uncorrelated variables (principal components) was the primary task of using the PCA. The above code fragment commences the PCA process by supplying the required number of components, which here was thoroughly selected at 10 by the author. Initially, the main components were computed using the scaled training dataset. Then, these components were individually applied to both the training and testing datasets. This approach not only reduced the dimensionality, but also preserved critical data, hence optimising subsequent data analysis while preserving data integrity.

```
# SVM algorithm
model = SVC(probability=True)
# Train the model
model.fit(X_train_pca, y_train)
```

Further, the above code line depicts that an instance of SVM classifier was initialised with a provision for probability estimation. The aim of having the probability as 'true' was to calculate the AUC while evaluating the model at a later stage. The subsequent step trained the classifier using PCA training data (X_train_pca) and their corresponding labels (y_train), ultimately fostering the development of the model capable of distinguishing patterns within the data.

### 5.5.6 Random Forest

Random Forest (RF) is a effective ensemble ML technique that has been widely used because it is capable to circumvent the constraints of a single decision trees. It is essentially a collection of decision trees with enhanced predictive capabilities. By combining the results of numerous decision trees, this algorithm achieves greater predicted accuracy and more stability (McFall et al., 2023).

At the core of the RF is the concept of bootstrap aggregation which is an ensemble pattern recognition method used in ML to improve the precision and resilience of prediction models, often known as bagging. By creating several bootstrap samples from the training data and training individual decision trees on these samples, the classifier provides variety among the constituent trees, hence minimising overfitting and increasing generalisation (McFall et al., 2023). In addition, the depth of each tree can be restricted, limiting the possibility of capturing data noise and advocating the model's simplicity.

### *Development*

Here, proceeding in the similar form as previous models, the author begun the development of the algorithm in python. Again, RFE feature extraction technique was applied as used in Logistic Regression model and the model was trained on the selected features.

```
# Create the RF model
model = RandomForestClassifier(max_depth=10)
# Train the model
model.fit(X_train_selected, y_train)
```

In the above stated code, an RF classifier was instantiated with a maximum depth of 10. The parameter named 'max depth' restricted the complexity (depth) of each decision tree, hence regulating the model's complexity and maybe improving its generalisation capabilities. The 'fit' approach used here trained the classifier on the feature engineered training data denoted by X_train_selected and y_train. The 'fit' function uses the algorithm to create decision trees repeatedly based on the input features and their related target labels. During the training phase, each tree is designed to best capture the underlying patterns in the data, while simultaneously incorporating randomisation to improve variety and prevent overfitting (McFall et al., 2023). By iteratively modifying the tree parameters and integrating their outputs, the model effectively learned the correlations between the data features and the 'PC' labels, resulting in a robust and generalisable classifier that was able to make fairly good predictions on unseen data.

### 5.5.7 Binary PSO & Random Forest

In an attempt to boost the Random Forest model's performance, the author redeveloped the RF algorithm using a distinctive *bio-inspired* feature selection technique called Particle Swarm Optimization (PSO).

PSO is a computational optimisation algorithm inspired by the flocking behaviour of birds. PSO is based on the concept of digitally simulating the flocking (of birds) in order to find an optimal solution in a multidimensional search area. Each particle represents a potential solution and so represents the solution (or feature subset in this case). These particles traverse the search space to locate the ideal solution based on their own and their neighbours' experiences (Pasha and Latha, 2020). In the background of PSO, a fitness function is defined to evaluate the quality of a feature subset.

#### Development

```
# Initialize the optimizer
optimizer = BinaryPSO(n_particles=num_particles, dimensions=X.shape[1],
options=options)
num_iterations = 50
```

The PSO optimiser was initialised using the BinaryPSO class from the PSO library of python. It was configured with the previously defined parameters and selections. The dimensions parameter here represents the number of features in the dataset. The algorithm runs for a specified number of iterations which here was restricted to 50 mainly because of the small dataset.

```
# Run PSO to select the best feature subset
best_cost, best_features = optimiser.optimise(evaluate_fitness,
iters=num_iterations)
```

Through the iterations it found the best feature subset that produced the best accuracy score, calculated by the 'evaluate fitness' function. PSO's optimise function updated the parameters recursively to refine the search for the optimal subset.

At the conclusion of PSO, the best feature subset determined based on the accuracy score was stored in the ‘best features’ variable.

```
# train and fit the final model
final_RF_model = RandomForestClassifier(max_depth=5)
final_RF_model.fit(X_train.iloc[:, best_features], y_train)
```

Finally, the RF model was trained with a max depth of 5 using the best feature subset found by the PSO algorithm. This final trained model was intended to be more accurate and efficient due to the selection of relevant features.

## 5.6 Model Optimisation

Optimisation is fundamental to ML model development which is achieved by tweaking the hyperparameters. Therefore, after the first successful run of the developed ML models and making a rough judgment of the accuracy achieved, the author optimised the models first by adding the optimal hyperparameters or removing them where necessary and further by tuning them. Some specific techniques were employed to remove the overfitting of the models wherever needed. As the optimisation was distinct for all the models, they are discussed separately.

***Decision Trees*** can be optimised by increasing or decreasing the depth hyperparameter. And after numerous trials, the author found the optimal value for depth constraint as 5 (*max_depth*=5).

***MLP*** was initially trained with hidden layer sizes as (250, 200) which was found to be highly complex for the small dataset used and was leading to overfitting. Moreover, the number of iterations (max_iter) was initialised at 200 but the optimisation couldn’t be achieved with these iterations. So, hidden layer sizes were reduced considerably to be fixed at (50, 30) and the ‘iterations’ hyperparameter was steadily increased and ultimately set at 500 for the training of the model to be completed optimally.

***KNN*** was initially trained with 10 neighbours (*k*). Later, to find the optimal value of *k*, the author employed cross-validation and experimented with different values of *k* and evaluated the model's performance using 5-fold cross-validation. The *k* value of *11* yielded the highest cross-validation accuracy which was eventually selected as the optimal *k* for the dataset.

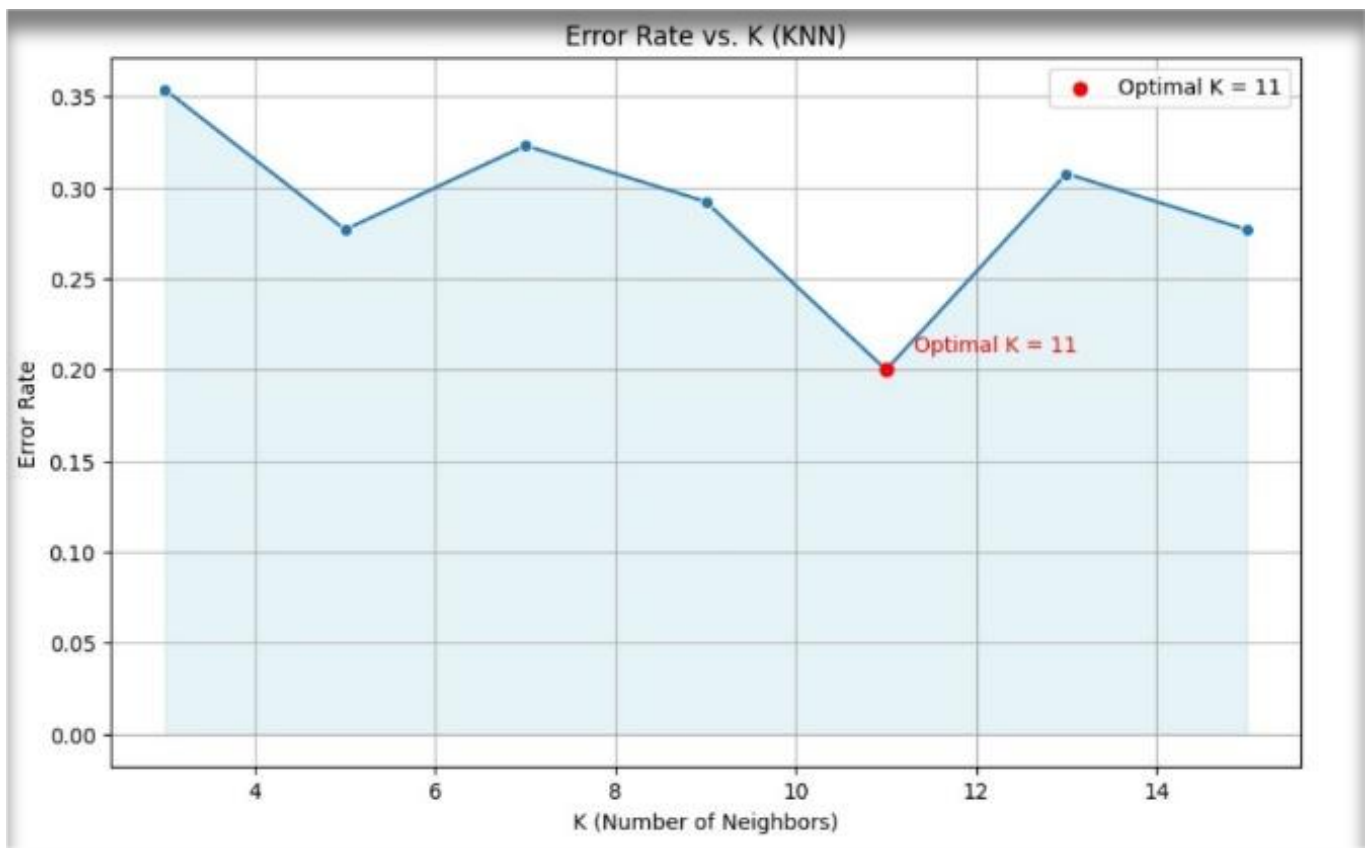

Fig 5.11 Optimal *k*-value for KNN

***Logistic Regression*** and ***SVM*** were essentially optimised using feature engineering which has already been discussed in development section of this report.

Although ***Random Forest*** was also optimised using feature engineering, but its development had to be controlled by the adding the depth constraint and ultimately setting it to 10 (*max_depth*=10).

While developing the ***Random Forest*** with ***PSO,*** the depth constraint set at 5 (*max_depth*=5) produced the appropriate results.

# Chapter – 6 EVALUATION AND RESULTS

At this stage, the author examined the performance of the trained models. The assessment was the most important component of the study because it revealed the worth of all the previous efforts. But, before delving into the details of the models' assessment, the author has described the metrics used and the ambit in which they were applied.

## 6.1 Evaluation Metrics

The evaluation of a trained ML model's performance is based on its capability and efficiency of making accurate predictions on unseen data. This unseen data is often called testing data which was reserved for evaluation purposes.

```
# Model evaluation
y_pred = model.predict(X_test)
acc = accuracy_score(y_test, y_pred)
```

In the above-mentioned code, the variable X_test represents the feature inputs of the testing dataset and the model's predictions are stored in the variable y_pred.

Thus, to quantify the performance of the models, some commonly used metrics were used including classification accuracy or accuracy, precision, sensitivity, specificity, cross-validation and area under the curve. Here, the author has presented the mathematical principles as well as the programming provisions for evaluation metrics implemented during the study.

***Accuracy*** is the ratio of the accurate predictions and the total number of predictions made of the ML model (Mishra, 2020).

$$\text{accuracy} = \frac{\text{correct predictions}}{\text{all predictions}}$$

The accuracy of model was calculated separately for its training and testing aspects. This was done to gauge the model for overfitting while training. A model with greater accuracy proves its better performance but it can be misleading when dealing with imbalanced dataset such as the one used in this study.

```
# Calculate training and test accuracies
y_train_pred = model.predict(X_train)
y_pred = model.predict(X_test)

train_acc = accuracy_score(y_train, y_train_pred)
test_acc = accuracy_score(y_test, y_pred)
```

***Precision*** is the ratio of appropriate instances (true positives) among all instances projected to belong to a particular class (Mishra, 2020).

$$\text{precision} = \frac{\text{true positives}}{\text{true positives} + \text{false positives}}$$

```
#Calculate precision
prec = precision_score(y_test, y_pred)
```

High precision is desired when minimising the false positives is crucial. It is mostly used when the getting false positives is not acceptable.

For the calculation of sensitivity and specificity, all the correct and incorrect outputs from models had to be calculated.

```
#parameter calculations
true_neg = np.sum((y_pred == 0) & (y_test == 0))
false_pos = np.sum((y_pred == 1) & (y_test == 0))
false_neg = np.sum((y_pred == 0) & (y_test == 1))
true_pos = np.sum((y_pred == 1) & (y_test == 1))
```

***Sensitivity(Recall)*** is the fraction of cases predicted to belong to a class and the total number of cases that actually belong to that class (Mishra, 2020).

$$\text{recall} = \frac{\text{true positives}}{\text{true positives} + \text{false negatives}}$$

```
#sensitivity calculation
sens = true_pos / (true_pos + false_neg)
```

High sensitivity is good when minimising false negatives is important which necessarily ensures that actual positive cases are not missed. Sensitivity was significant in this study where a medical diagnostics was being explored where missing positives could prove costly.

***Specificity*** quantifies the true negative predictions out of all actual negative instances. It is produced as a ratio.

$$Specificity = \frac{TN}{TN + FP}$$

```
#specificity calculation
specif = true_neg / (true_neg + false_pos)
```

High specificity is good when minimising false positives is important, especially when the negative class is more critical.

***Area Under Curve (AUC)*** is ability of an ML model to distinguish between binary classes of a problem scope. To calculate the AUC, model's true positive rate is plotted against its false positive rate and area covered by the plotted graph is determined to get the AUC score (Mishra, 2020). By producing this curve, the author has visually evaluated the performance of the classifier across a range of decision thresholds.

```
# Model evaluation
y_pred_prob = model.predict_proba(X_test_selected)[:, 1]  # Get the
probability of the positive class (class 1)

# Calculate ROC curve
fpr, tpr, thresholds = roc_curve(y_test, y_pred_prob)

# Calculate the area under the ROC curve (AUC)
roc_auc = auc(fpr, tpr)
```

High AUC indicates a classifier's effectiveness at differentiating positive cases from negative cases which is crucial for this study to discern PD patients from healthy controls.

***Cross-validation*** is widely used to assess an ML model's performance on unseen data and to mitigate potential biases caused by the train-test split. The process begins by dividing the dataset into *k* subsets, or folds. The model is then trained and evaluated *k* times, each time using a different fold as the test set and the remaining four folds as the training set. This allows the model to be tested on various data samples, which helps to get a more robust estimation of its generalisation performance (Spetsieris et al., 2010).

```
#cross-validation
cv_scores = cross_val_score(model, X_train, y_train, cv=5)  # Perform
5-fold cv

# average accuracy across all folds
cv_accuracy = cv_scores.mean()
```

Cross-validation is primarily utilised to ensure the model doesn't overfit the training data.

## 6.2 Results

The author has presented the results of each model in the form of a table. An interpretation of the results is also provided with the outcome tables.

### *Decision Trees*

Table – 6.1

| Metric | Value |
|---|---|
| Cross-validation scores | [0.6154, 0.6154, 0.7115, 0.6923, 0.6923] |
| CV Scores Standard deviation | 0.0463 |
| Cross-validation accuracy | 0.6654 |
| Model Accuracy | 0.7538 |
| Training Accuracy | 0.9731 |
| Testing Accuracy | 0.7538 |
| Precision | 0.7660 |
| Sensitivity | 0.8780 |
| Specificity | 0.5417 |
| AUC | 0.75 |

### *Interpretation*

While the model's cross-validation accuracy of 66.54% suggests that it achieved a moderate level of generalisation, its accuracy of 75.38% on overall dataset demonstrates its ability to classify a good amount of the instances correctly. The relatively low standard deviation of CV scores (0.0583) suggests consistent performance across folds. Particular to PD diagnostics, the precision of 76.60% signifies the proportion of PD patients that were predicted correctly among all positive predictions, which was important for minimising false positives.

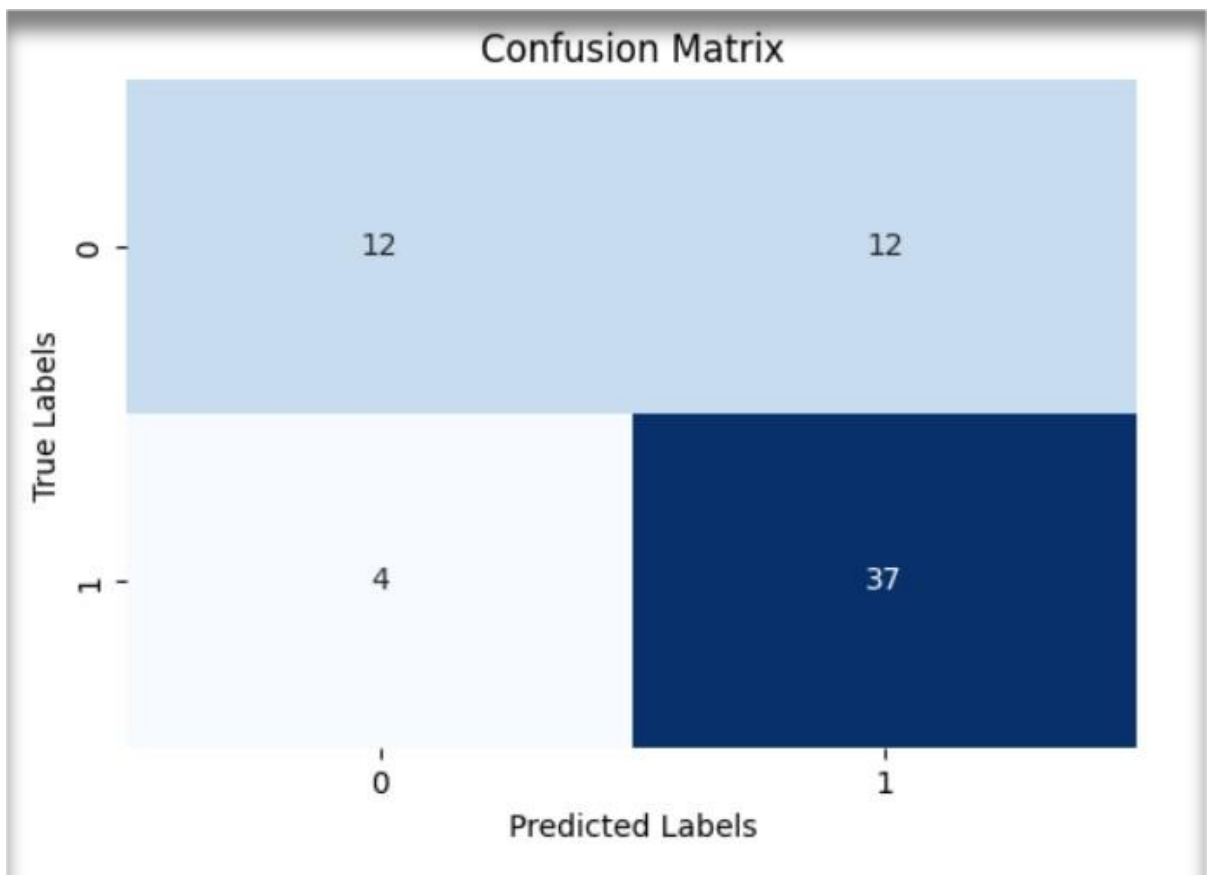

| TN = 12 | FP = 12 |
|---|---|
| FN = 4 | TP = 37 |

Fig 6.1 Confusion Matrix – Decision Tree

The sensitivity value of 87.80% reflects that the model was able to accurately identify most of the true positive cases, crucial for capturing actual PD instances where the cost of having false negative can be detrimental. The lower specificity at 54.17% suggests the model's challenge in correctly identifying true negative cases, which could lead to higher false positive rates in real world scenarios.

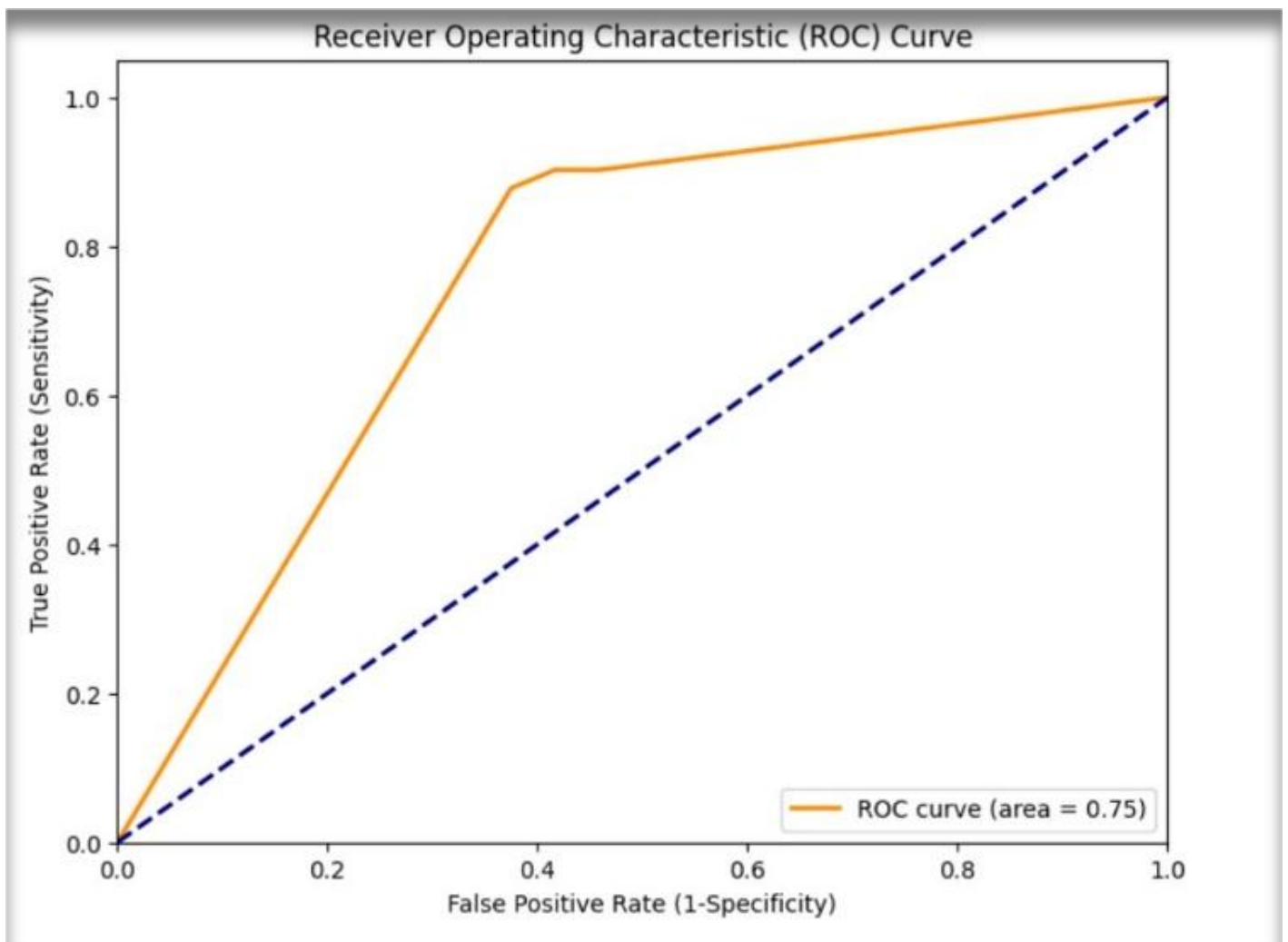

Fig 6.2 AUC – Decision Tree

Generally, an AUC > 0.5 is considered to have an acceptable level of discriminative ability, with higher values indicating better performance (Hicks et al., 2022). So, AUC value of 0.75 in the ROC curve represents the model's descent ability to discriminate between positive and negative cases.

## ***MLP Classifier***

Table – 6.2

| Metric | Value |
|---|---|
| Cross-validation scores | [0.7115, 0.6731, 0.6731, 0.8077, 0.6731] |
| CV Scores Standard deviation | 0.0583 |
| Cross-validation accuracy | 0.7077 |
| Model Accuracy | 0.6769 |
| Training Accuracy | 0.9885 |
| Testing Accuracy | 0.6769 |
| Precision | 0.7083 |
| Sensitivity | 0.8293 |
| Specificity | 0.4167 |
| AUC | 0.79 |

### *Interpretation*

The model's cross-validation accuracy (70.77%) indicates its reasonable performance in generalising the test data. The relatively low standard deviation of CV scores (0.0583) suggests consistent performance across folds. MLP model’s accuracy (67.69%) and the precision (70.83%) indicates a decent ability to correctly classify positive cases.

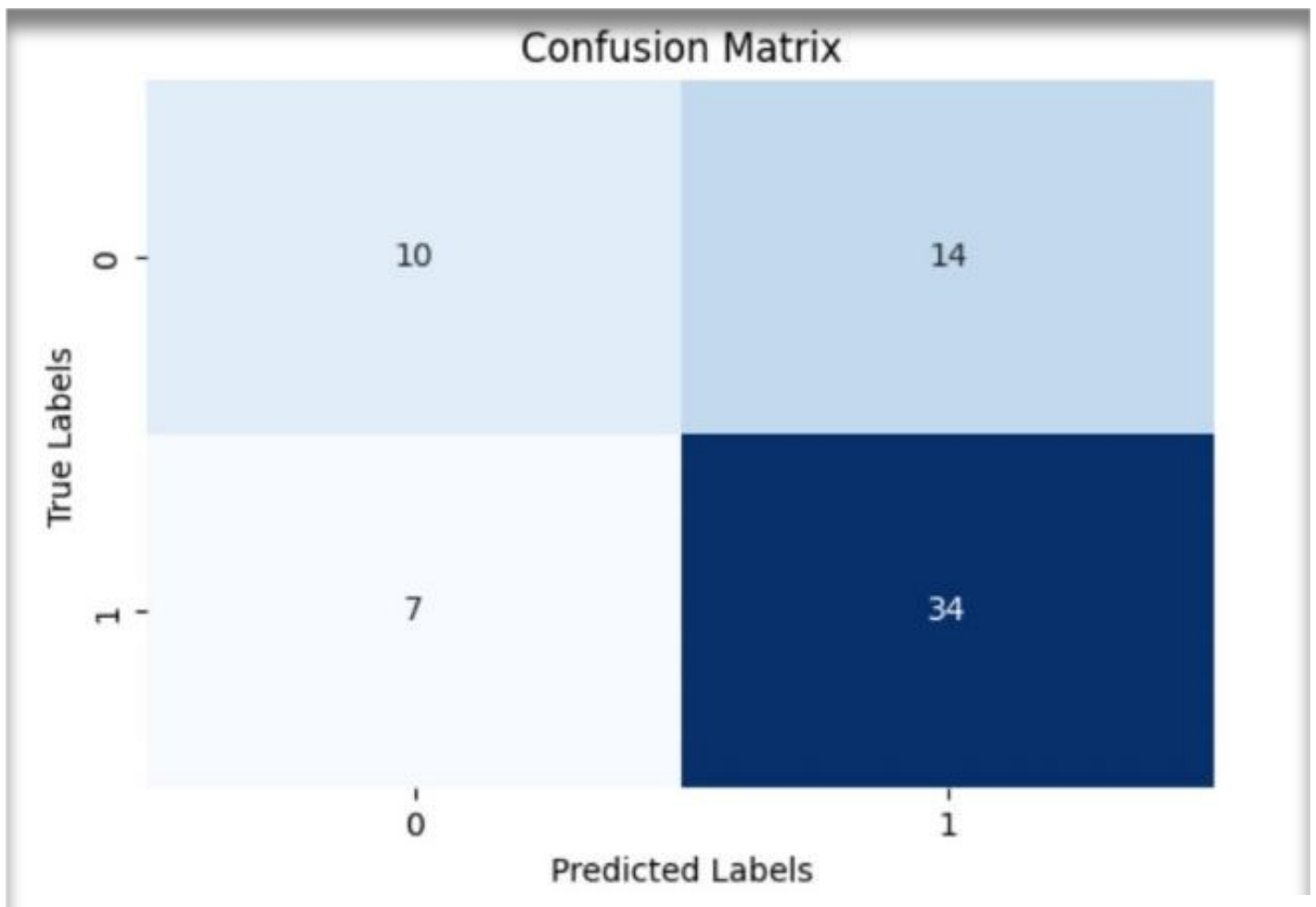

| | |
|---|---|
| TN = 1 | FP = 14 |
| FN = 7 | TP = 34 |

Fig 6.3 Confusion Matrix - MLP

While sensitivity (82.93%) highlights its proficiency in identifying actual positive instances, low specificity (41.67%) indicates a struggle in distinguishing true negative cases.

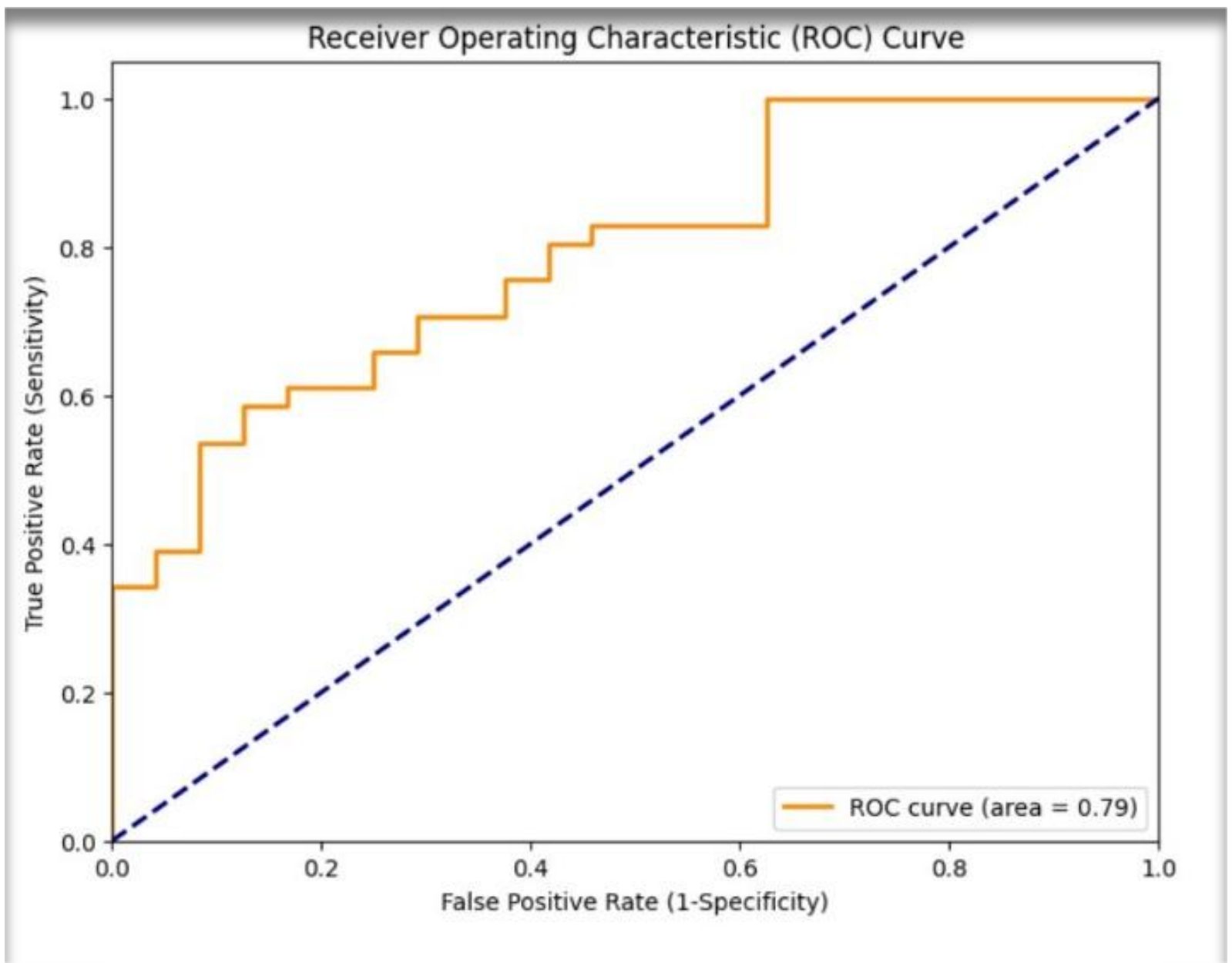

Fig 6.4 AUC - MLP

The overall model performance is underscored by the AUC of 0.79, indicating a good ability to distinguish between classes.

## *KNN*

Table – 6.3

| Metric | Value |
|---|---|
| Cross-validation scores | [0.7500, 0.7308, 0.6538, 0.6923, 0.7885] |
| CV Scores Standard deviation | 0.0520 |
| Cross-validation accuracy | 0.7231 |
| Model Accuracy | 0.7231 |
| Training Accuracy | 0.7731 |
| Testing Accuracy | 0.7231 |
| Precision | 0.7170 |
| Sensitivity | 0.9268 |
| Specificity | 0.3750 |
| AUC | 0.79 |

### *Interpretation*

KNN model demonstrated a consistent and balanced performance, as indicated by the cross-validation accuracy of 72.31% and a low standard deviation of 0.0520 in the CV scores, suggesting stable and reliable results. The model's overall accuracy of 72.31% on the testing dataset aligns closely with its performance during training (77.31%), underscoring its competency to generalise the unseen data

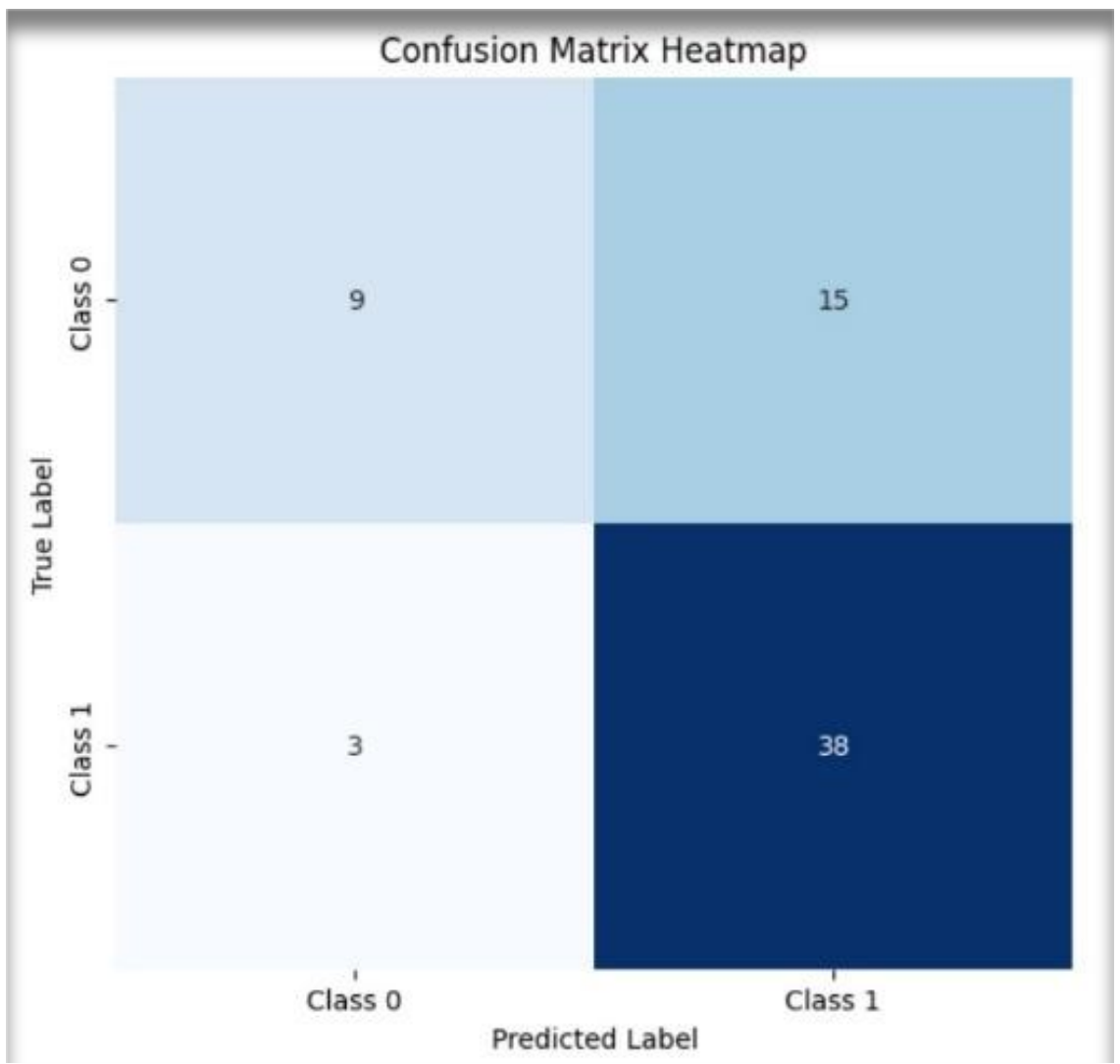

| | |
|---|---|
| TN = 9 | FP = 15 |
| FN = 3 | TP = 38 |

Fig 6.5 Confusion Matrix – KNN

Additionally, the precision of 0.7170 implies a reasonable balance between correctly identifying positive cases and avoiding false positives. The high sensitivity value of 0.9268 reflects the model's proficiency in identifying true positives among actual positive instances. However, the lower specificity of 0.3750 signifies a challenge in correctly identifying true negatives, which could be an area for improvement.

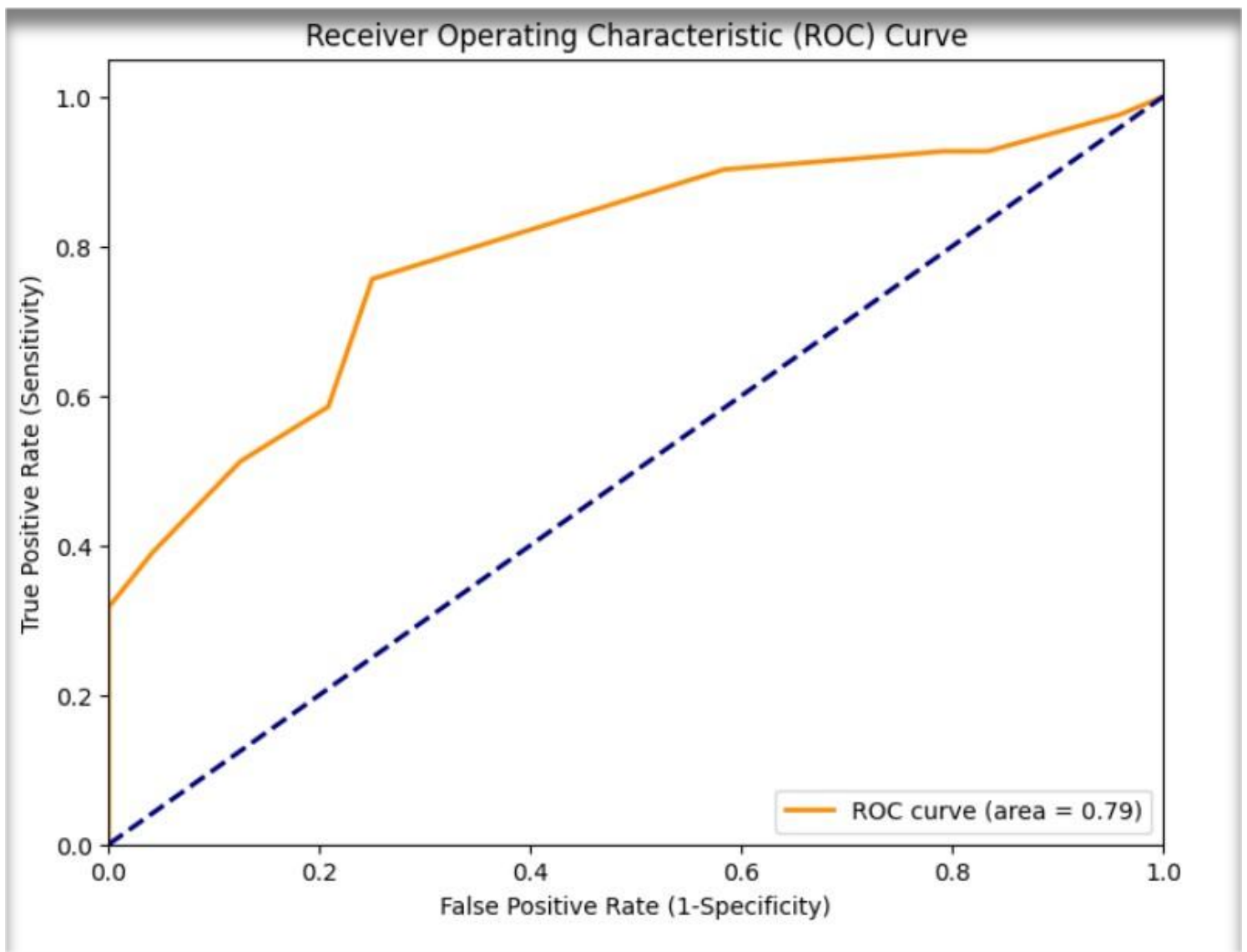

Fig 6.6 AUC – KNN

The achieved AUC of 0.79 suggests a great ability of the model to distinguish between positive and negative instances.

## ***Logistic Regression***

Table – 6.4

| Metric | Value |
|---|---|
| Cross-validation scores | [0.7692, 0.7692, 0.6731, 0.6923, 0.7885] |
| CV Scores Standard deviation | 0.0520 |
| Cross-validation accuracy | 0.7385 |
| Model Accuracy | 0.7231 |
| Training Accuracy | 0.7346 |
| Testing Accuracy | 0.7231 |
| Precision | 0.7091 |
| Sensitivity | 0.9512 |
| Specificity | 0.3333 |
| AUC | 0.75 |

## *Interpretation*

From the results given in the above table, the author deduced the interpretation that a standard deviation of 0.0520 in cross-validation scores signifies model's fairly stable performance across different folds. And the model had a substantial ability to generalise (73.85%) on the testing data. The overall model accuracy (73.31%) suggests that the model was competent enough to classifying a decent number of instances correctly.

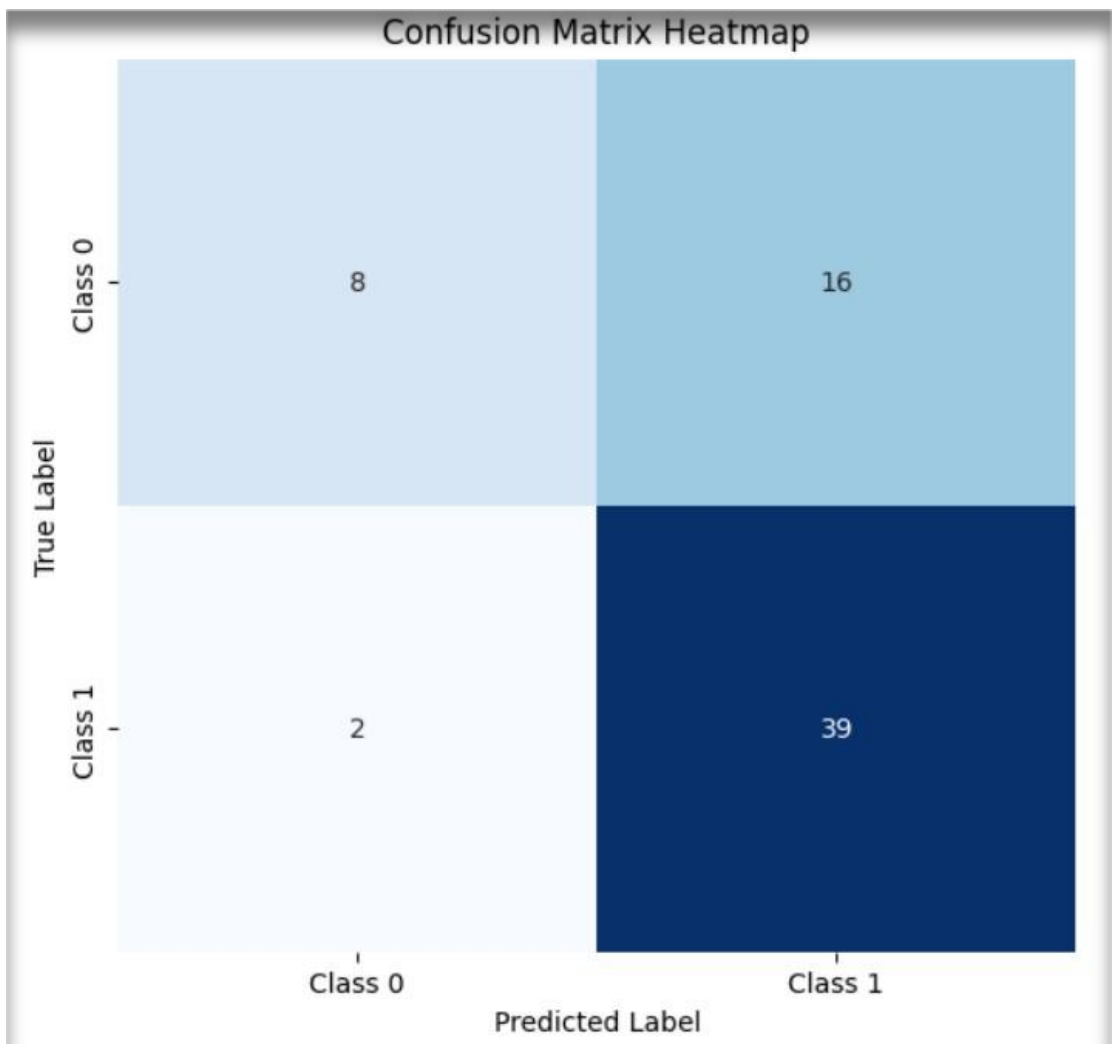

| | |
|---|---|
| TN = 8 | FP = 16 |
| FN = 2 | TP = 39 |

Fig 6.7 Confusion Matrix – Logistic Regression

Precision (70.91%) reflects the model was able to identify most of the positive PD cases correctly among all cases that were predicted as positive, which is vital in medical diagnosis to minimise the false positives. Exceptional sensitivity value (95.12%) of the model characterises the model's capacity to identify actual positives, an extremely critical aspect in detecting medical conditions to avoid false negatives. The specificity (33.33%) reveals that the model almost failed in accurately identifying true negatives, highlighting room for improvement in distinguishing non-PD cases.

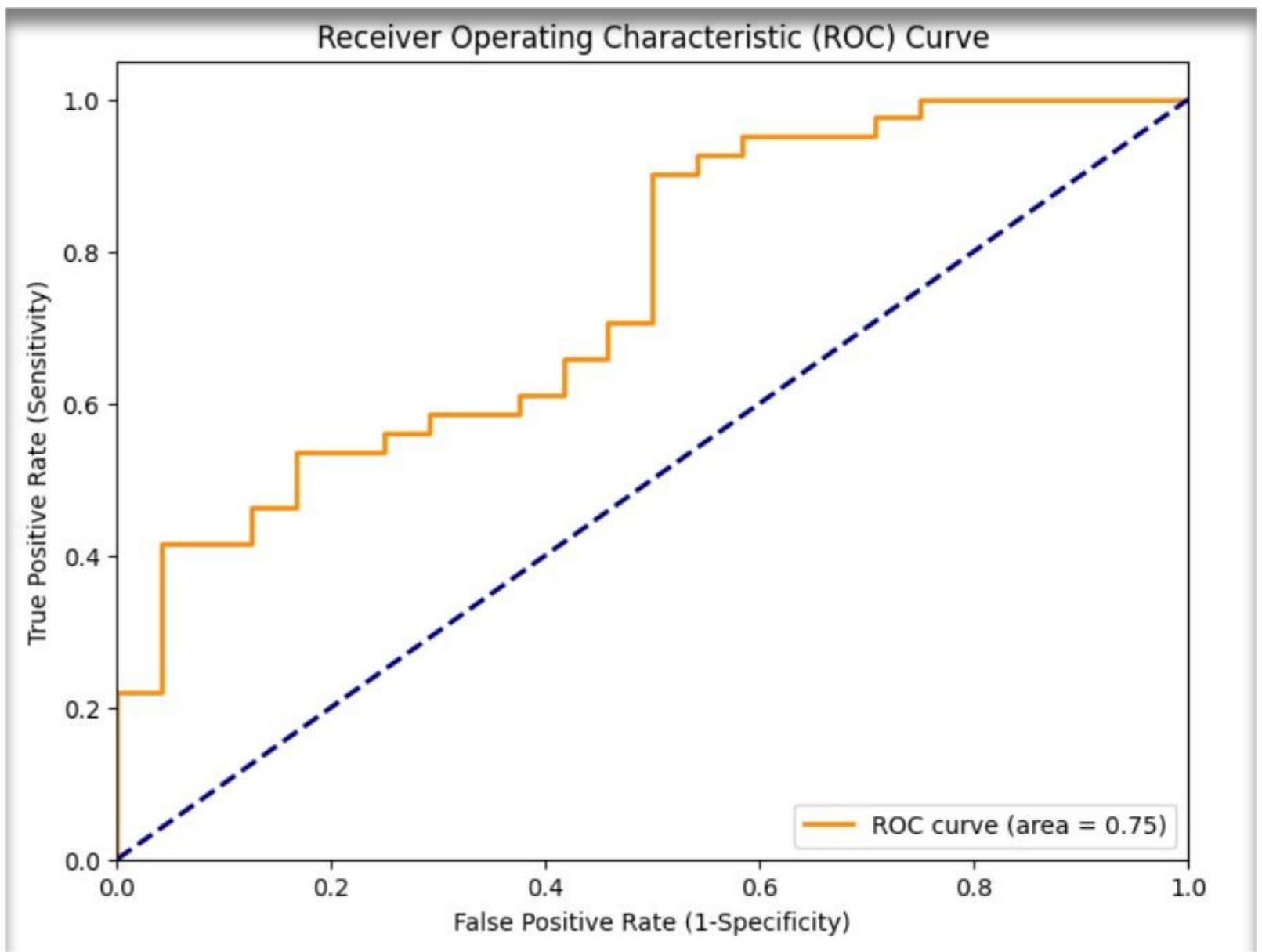

Fig 6.8 AUC – Logistic Regression

From the AUC in fig 6.8, the author understood that the model showed a reasonable (75.00%) level of discrimination ability which means the model can tell between positive and negative cases with moderately high accuracy. Importantly, the consistency between training accuracy (73.46%) and testing accuracy (72.31%) suggests a fairly aligned model. In conclusion, the model demonstrates strong potential in identifying positive cases with high sensitivity, but there is a need to enhance specificity and overall discriminative power for accurate medical predictions.

## ***<u>SVM</u>***

Table – 6.5

| Metric | Value |
|---|---|
| Cross-validation scores | [0.7308, 0.7500, 0.6923, 0.7692, 0.7692] |
| CV Scores Standard deviation | 0.0322 |
| Cross-validation accuracy | 0.7423 |
| Model Accuracy | 0.7231 |
| Training Accuracy | 0.7692 |
| Testing Accuracy | 0.7231 |
| Precision | 0.7018 |
| Sensitivity | 0.9756 |
| Specificity | 0.2917 |
| AUC | 0.68 |

### *<u>Interpretation</u>*

The author deduces from the results that the model performed fairly well on generalising the new data, given its cross-validation accuracy of 74.23%. The training accuracy of 76.92% signifies that the model fit fine on the training data while the testing accuracy of 72.31% shows a comparatively lower competence which can be a hint towards overfitting.

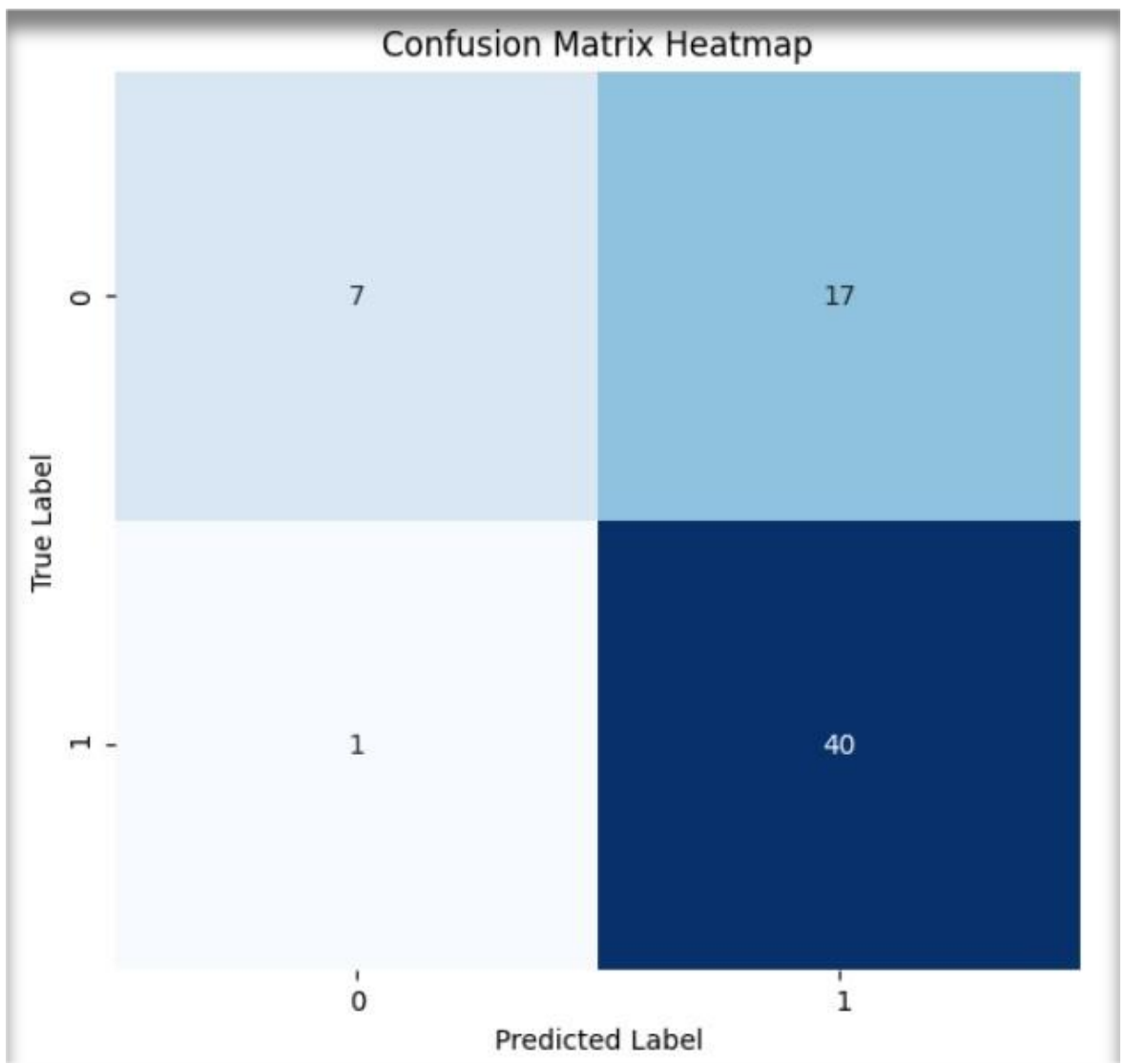

| | |
|---|---|
| TN = 7 | FP = 17 |
| FN = 1 | TP = 40 |

Fig 6.9 Confusion Matrix – SVM

Precision at 70.18% showcases the proportion of correctly predicted positive cases among all predicted positives. The remarkably high sensitivity of 97.56% implies a small rate of false negatives, which is extremely crucial in the medical context to avoid missing actual positive cases. However, the low specificity of 29.17% suggests a higher rate of false positives, indicating that the model was labelling too many non-PD as PD patients.

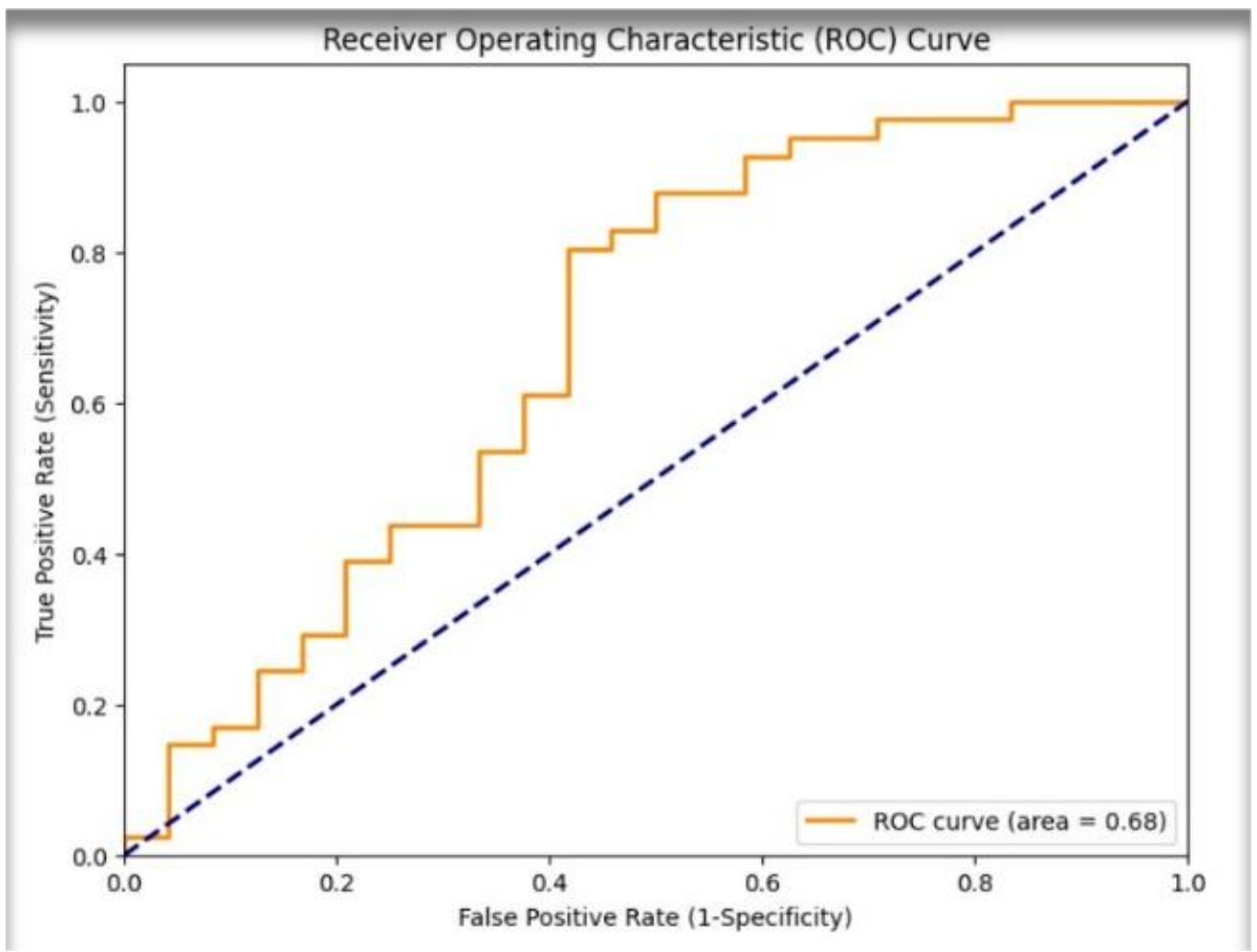

Fig 6.10 AUC – SVM

The AUC value of 0.68, which is above 0.5 but not extremely high, implies that the model has some discriminatory power in distinguishing between positive and negative cases.

## ***Random Forest***

Table – 6.6

| Metric | Value |
|---|---|
| Cross-validation scores | [0.6923, 0.6923, 0.7692, 0.7308, 0.7500] |
| CV Scores Standard deviation | 0.0344 |
| Cross-validation accuracy | 0.7269 |
| Model Accuracy | 0.7538 |
| Training Accuracy | 1.0000 |
| Testing Accuracy | 0.7538 |
| Precision | 0.7551 |
| Sensitivity | 0.9024 |
| Specificity | 0.5000 |
| AUC | 0.77 |

## *Interpretation*

Interestingly, from the training accuracy (100%) of the RF classifier, it was clear that it had memorised the training data, which rose concerns about generalisation capabilities of the model to new data. Even after several attempts to tweak, add and remove the hyperparameters, the model went into overfitting. So, the author went even deeper into optimising the algorithm by the using the bio-inspired optimisation algorithms (Random Forest & PSO).

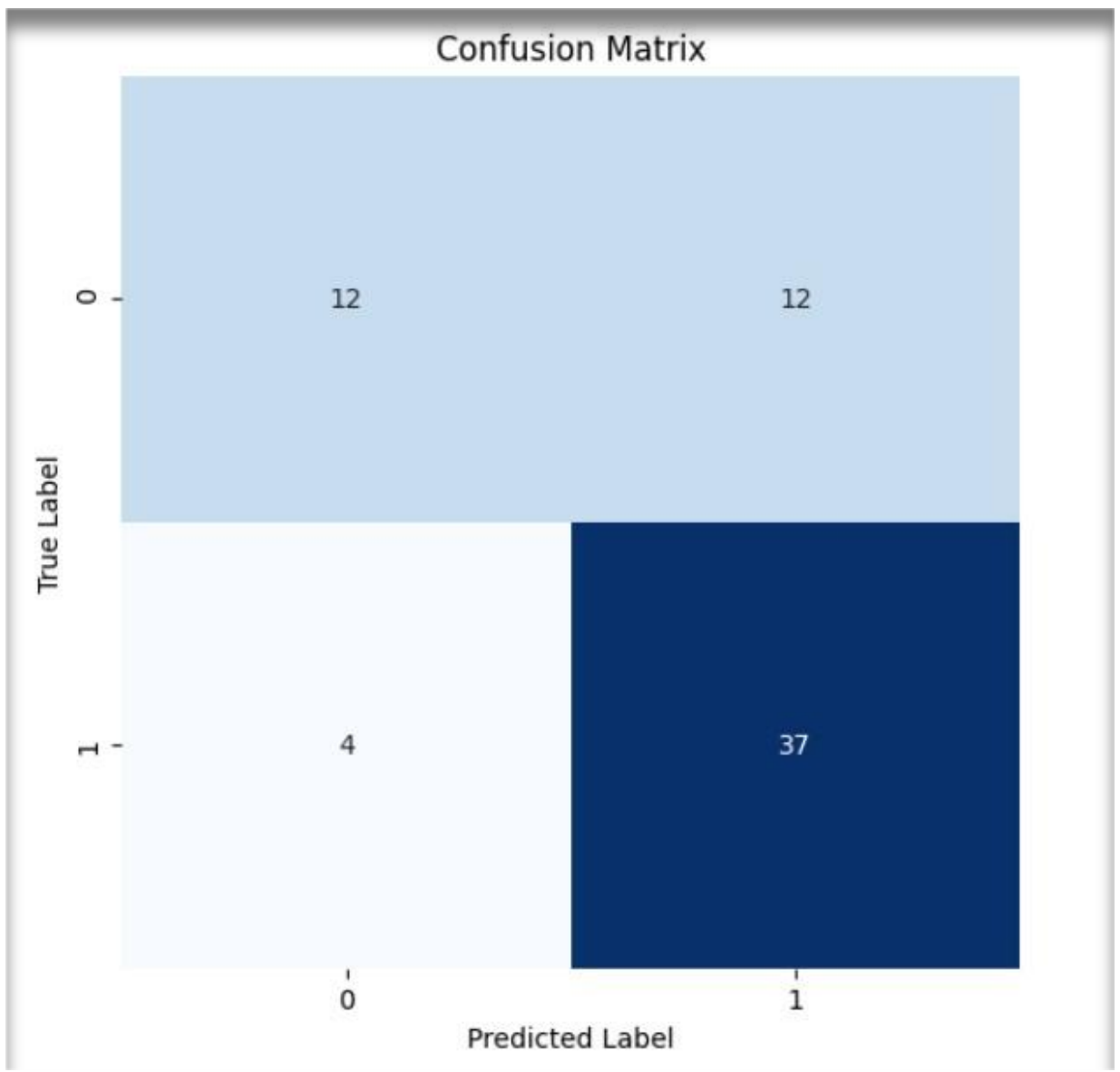

| | |
|---|---|
| TN = 12 | FP = 12 |
| FN = 4 | TP = 37 |

Fig 6.11 Confusion Matrix - Random Forest

The model's overall accuracy (75.38%) on unseen data reflects its proficiency in making correct predictions. The precision (75.51%) highlights the proportion of correctly identified positive PD cases among all cases predicted as positive. High sensitivity (90.24%) indicates the model's effectiveness in identifying actual positive cases, which is particularly important for not missing potential PD patients who require medical attention. The specificity of 0.5000 implies that the model struggled in correctly identifying negative cases, pointing to a potential need for improvement in distinguishing healthy controls.

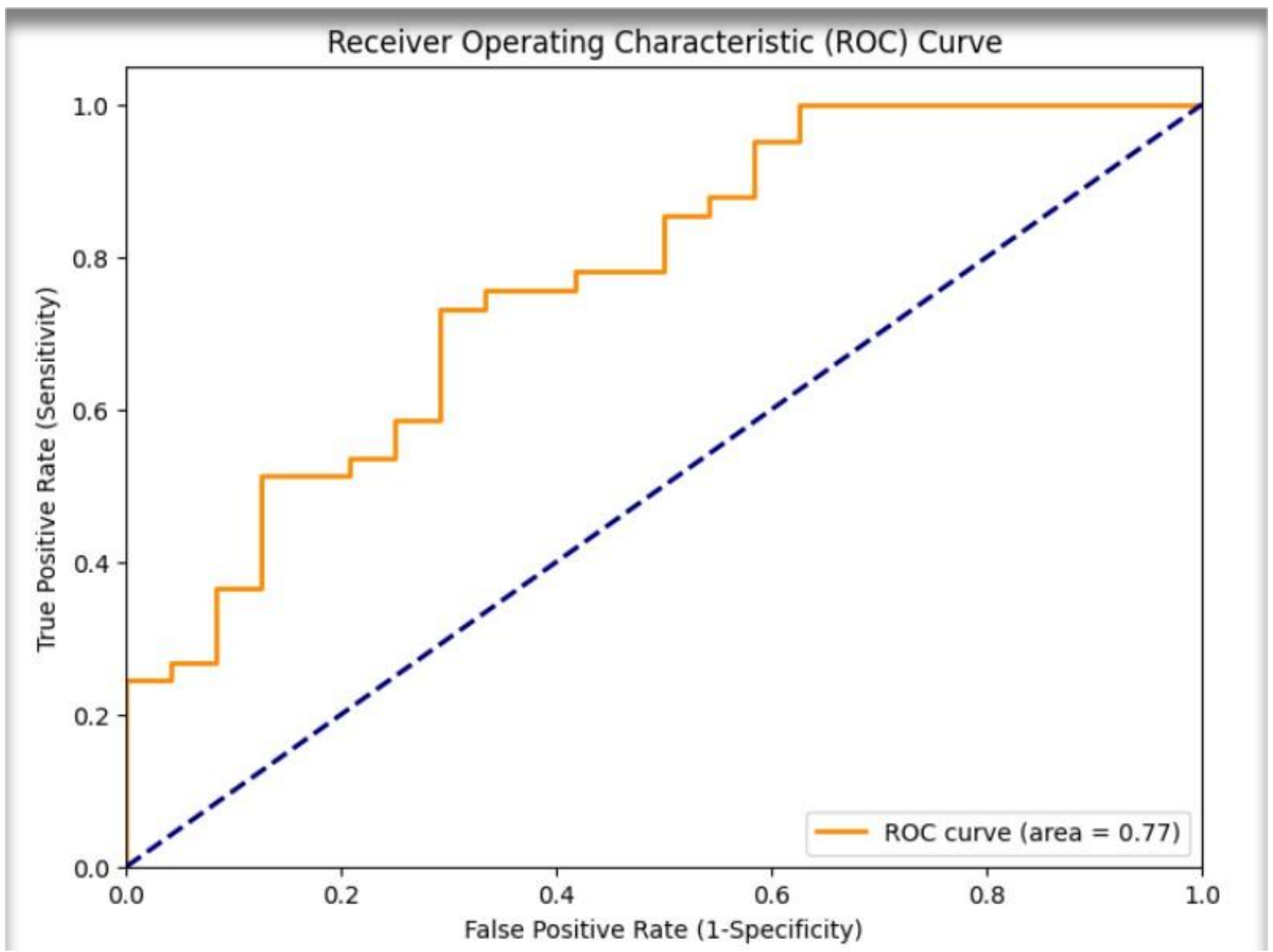

Fig 6.12 AUC - Random Forest

The AUC score of (77%) suggests a reasonably good ability of the model to discriminate between positive and negative cases. Although, the model did provide average outputs while classifying the PD cases, but an *overfit model* is often considered unreliable.

## ***Random Forest & PSO***

Table – 6.7

| Metric | Value |
|---|---|
| Cross-validation scores | [0.5538, 0.7692, 0.6615, 0.7385, 0.7077] |
| CV Scores Standard deviation | 0.0840 |
| Cross-validation accuracy | 0.6862 |
| Model Accuracy | 0.6308 |
| Training Accuracy | 0.7000 |
| Testing Accuracy | 0.6308 |
| Precision | 0.6308 |
| Sensitivity | 1.0000 |
| Specificity | 0.0000 |
| AUC | 0.47 |

### *Interpretation*

Although adopting bio-inspired optimisation in ML is becoming increasingly popular in AI, the author's attempt to optimise the RF classifier using PSO proved unsuccessful. Firstly, the model's relatively higher cross-validation standard deviation of 0.0840 demonstrated it to be unstable across different test data folds. Also, the training accuracy of 0.7000 and the overall model accuracy of 0.6308 were indicative of the model's poor performance, considering the extent of the optimisation performed prior to fitting the model to the training data from the feature subset.

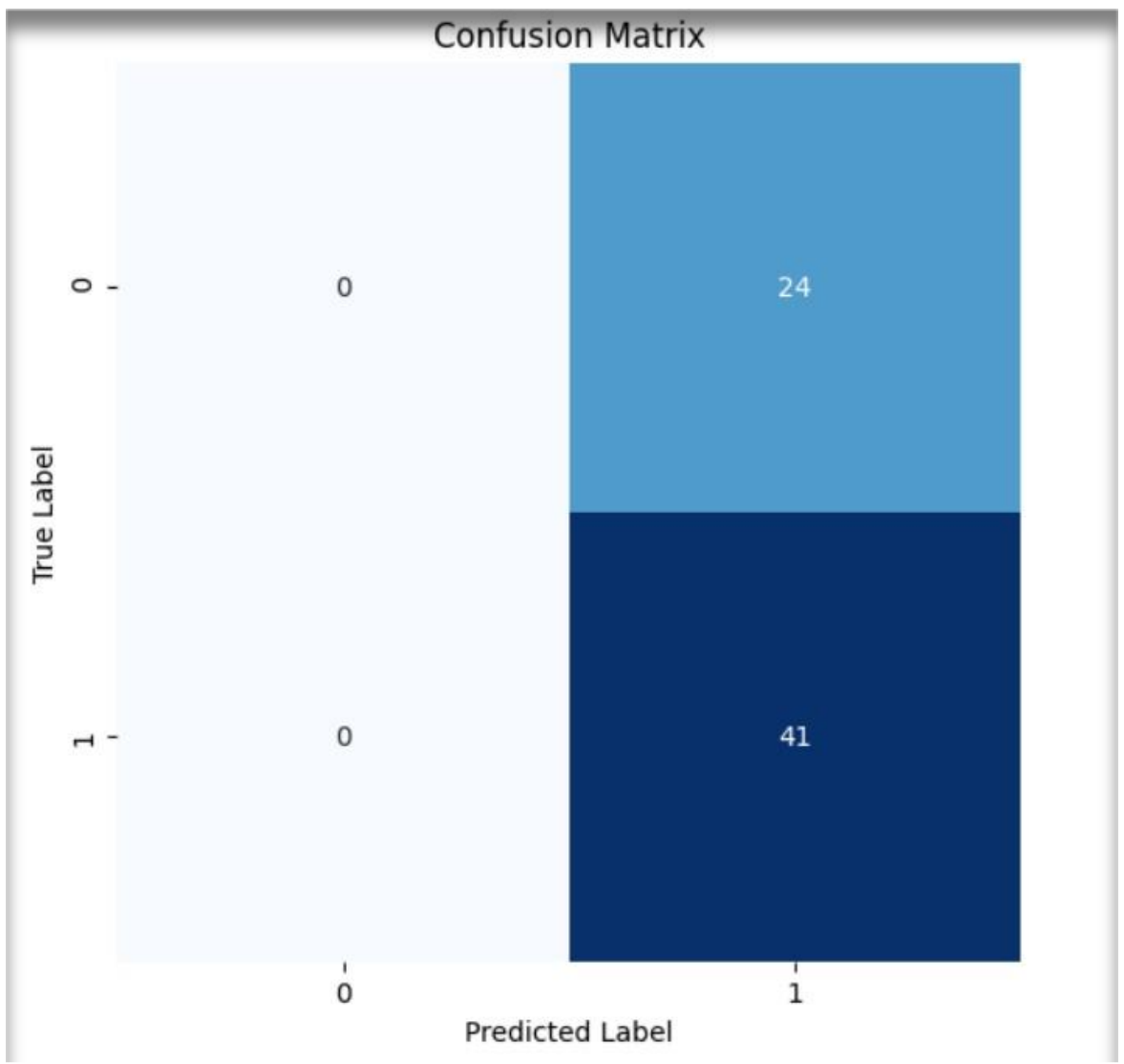

| | |
|---|---|
| TN = 0 | FP = 24 |
| FN = 0 | TP = 41 |

Fig 6.13 Confusion Matrix – Random Forest & PSO

Notably, the models sensitivity and specificity of 1.0000 & 0.0000 respectively, hinted towards possible *lazy learning*; where the model does not understand or learnt the data but predicts every instance as positive.

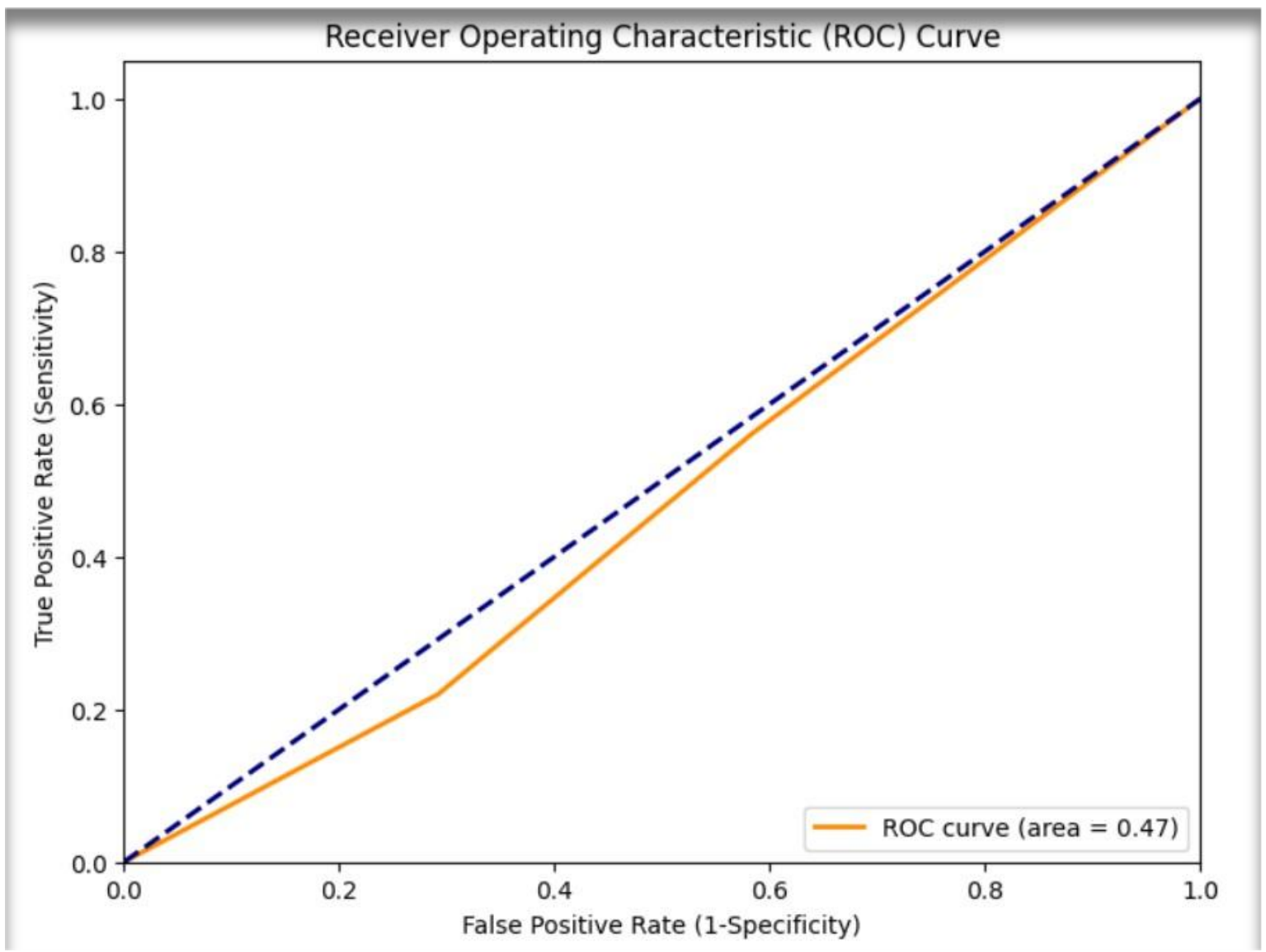

Fig 6.14 AUC – Random Forest & PSO

An AUC value of 0.47 which is less than 0.5 suggests the model's capacity to classify positive and negative cases was limited. Additionally, an ROC graph below the diagonal line often denotes that the model was making inverse predictions related to the true class labels. In other words, it was systematically getting predictions wrong, which possibly indicates some fundamental issues with the model.

# Chapter – 7 DISCUSSION AND FUTURE WORK

## 7.1 Comparison

| Model | Cross-validation accuracy | Model Accuracy | Training Accuracy | Testing Accuracy | Precision | Sensitivity | Specificity | AUC |
|---|---|---|---|---|---|---|---|---|
| Decision Tree Classifier() | 0.6654 | 0.7538 | 0.9731 | 0.7538 | 0.766 | 0.878 | 0.5417 | 0.75 |
| MLP Classifier() | 0.7077 | 0.6769 | 0.9885 | 0.6769 | 0.7083 | 0.8293 | 0.4167 | 0.79 |
| k-Neighbors Classifier() | 0.7231 | 0.7231 | 0.7731 | 0.7231 | 0.717 | 0.9268 | 0.375 | 0.79 |
| Logistic Regression() | 0.7385 | 0.7231 | 0.7346 | 0.7231 | 0.7091 | 0.9512 | 0.3333 | 0.75 |
| SVM() Classifier | 0.7423 | 0.7231 | 0.7692 | 0.7231 | 0.7018 | 0.9756 | 0.2917 | 0.68 |
| Random Forest Classifier() | 0.7692 | 0.7538 | 1 | 0.7538 | 0.7551 | 0.9024 | 0.5 | 0.77 |
| Random Forest Classifier() and PSO | 0.6862 | 0.6308 | 0.7 | 0.6308 | 0.6308 | 1 | 0 | 0.47 |

Table – 7.1 Evaluation Table

In terms of accuracy, ML models produced a score in the range of 66-76%. RF and MLP classifiers demonstrated a competitive performance with cross-validation accuracy around 72-73%, while Decision Tree and *k*-Neighbours achieved highest model accuracy of 72-76%. As the dataset was imbalanced, actual performance of the models was ascertained from their AUC scores with an average of 0.75 which was considered to be relatively good.

Conspicuously, RF classifier's performance decreased greatly when optimised with PSO. It leading to a perfect sensitivity and no specificity signifies that the classifier was not learning the data and was only poorly guessing the instances.

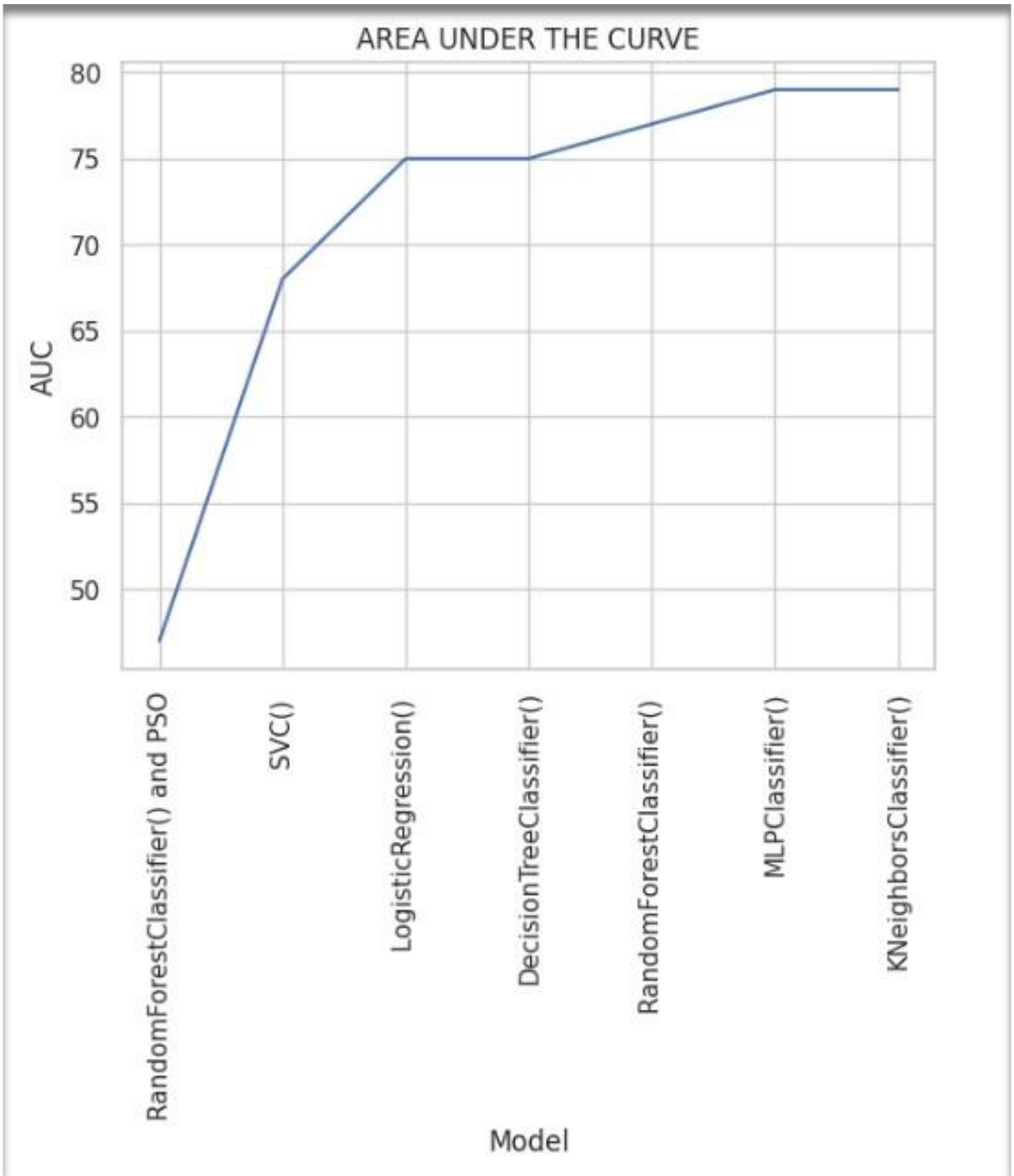

*Fig 7.1 Area Under the Curve*

Fig 7.1 shows that Decision Tree, MLP, and *k*-Nearest Neighbours, demonstrated solid performance around 0.75-0.79 AUC score, while Logistic Regression and Random Forest also exhibited competitive results. However, integrating Particle Swarm Optimization with Random Forest led to a notable drop in AUC to 0.47.

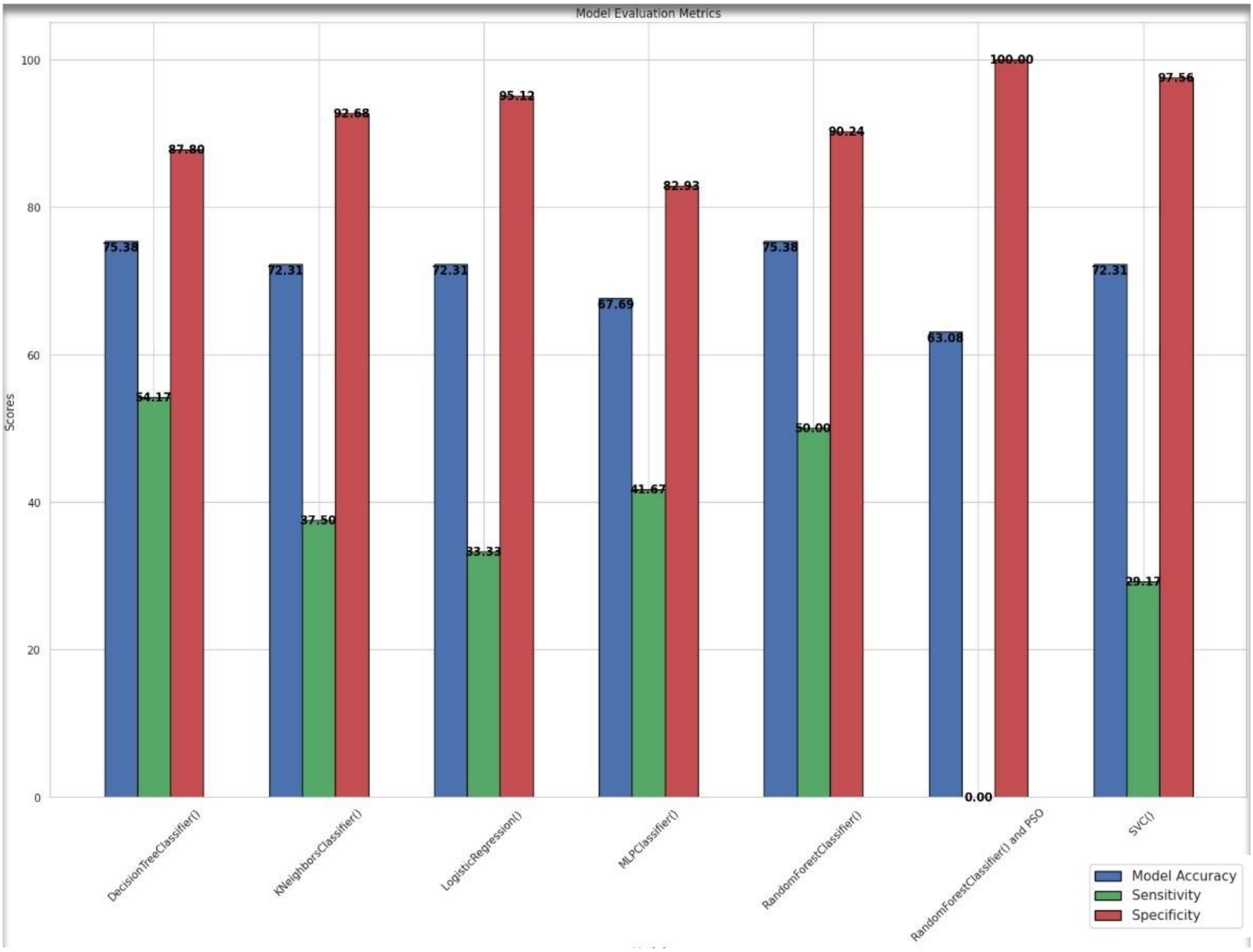

*Fig 7.2 Accuracy, Sensitivity & Specificity*

## 7.2 Conclusion

In conclusion, this dissertation delved into the realm of PD diagnosis utilising a comprehensive array of ML algorithms aiming to explore PD diagnosis by applying these advanced computational techniques.

Extensive testing and analysis revealed that the ML algorithms were fairly effective for categorising PD patients in the presented dataset. The results established that the algorithms showcased varying degrees of performance, with some showing better predicting ability than others. General PD case prediction accuracy was highest for the Decision Tree and RF classifiers, whereas all things considered, MLP was abortive. In addition, SVM achieved an excellent sensitivity, meaning it exhibited exceptional capacity to identify the true positive cases, which is crucial in the medical sector to minimise false positives. Furthermore, RF classifier exhibited a reasonable ability to capture complex relationships in the data whose accuracy was driven by high training accuracy (100%) which although hinted towards potential overfitting. Nonetheless, the optimisation of RF classifier with bioinspired PSO led to further reduction in its performance and displayed a possible case of lazy learning.

Additionally, the author witnessed a low specificity across the board which means all the models failed to recognise the non-PD cases (healthy controls). The possible reason could be the data having significantly lesser number of healthy instances than confirmed PD patients which is often referred to as imbalanced dataset. In such cases, the provided data is typically insufficient for adequate training of the models for the expected tasks.

Ultimately, the findings of this dissertation highlight the significance of employing a variety of ML algorithms and optimisation techniques to address such complicated medical diagnosis tasks. The combination of these methods (ensemble models) possibly will result in more precise and trustworthy non-clinical diagnosis of PD, which are crucial for early intervention and treatment of incurable neurodegenerative illnesses.

## 7.3 Limitations

This section discusses the limitations observed by the author while carrying out of this study –

- *Low sample size* – dataset had only 19 feature columns and 325 rows. The small data potentially impacted the correlation and pattern reading ability of the ML algorithms employed, leading to overfit and underfit predictive models. Affecting the overall purpose of this study.
- *Specificity issue* – training a model on an unbalanced data is considered as a substantial shortcoming in the field of ML. In the context of this study, the unbalanced data at hand underrepresented the non-PD cases which led to biased model predictions and reduced specificity in detecting the minority class.
- *Data augmentation* – not having used any data augmentation techniques, the author learned that it impeded the capacity of the models to predict the healthy controls which degraded the generalisability of the models.
- *Algorithm selection* – although the experimentation in this dissertation were carried out after exhaustive review of previous work in the area of applications of ML in the medical sector, the author in this dissertation while selecting the various ML algorithms could have overlooked some other potential models which might have yielded better outcomes.
- *Limited Exploration of Deep Neural Networks* – in this study, the author focused only on usage of traditional ML models for PD exploration, not tapping the potential of deep neural networks which are more advanced ML techniques capable of handling complex data.
- *Time constraint* – limited time available was another obstacle in further study of the ML models and their experimentation with the PD dataset.

## 7.4 Future Work

While indulging into further work for this study, the author would include the following –

- *Data size* – the author would collect or acquire a larger dataset with more features and data rows which will possibly enhance the yielding of the ML algorithms while reducing the shortcomings like overfitting.
- *Data balancing* – due to time restrictions the author could not employ the data balancing and augmentation techniques such as SMOTE which creates added synthetic samples during the model training for the minority class. This will address the low specificity issue of the models which needed attention.
- *Deep neural networks* – the author would employ deep networks such as Recurrent and convolutional neural network which are effective where time-series data such as patient health records are involved.
- *Explore latest techniques* – recent techniques like automated machine learning can be also explored for the further experimentation of PD. AutoML by Google Cloud, IBM AutoAI, etc provide the platforms for the automating various ML tasks.

In closing, this dissertation contributes to the growing body of research in utilising ML for medical diagnosis through a careful fusion of refined feature engineering, algorithmic sophistication, and rigorous validation, specifically in the context of Parkinson's disease. The insights gained from this study can guide further investigations and inspire the development of more refined and accurate diagnostic tools, ultimately leading to improved patient care and outcomes in the realm of neurodegenerative diseases.